\documentclass{article}

\usepackage{arxiv}
\usepackage[utf8]{inputenc} 
\usepackage[T1]{fontenc}    
\usepackage{hyperref}       
\usepackage{url}            
\usepackage{nicefrac}       
\usepackage{microtype}      

\usepackage{lipsum}         
\usepackage{graphicx}
\usepackage{natbib}
\usepackage{doi}

\usepackage{amsmath,amsfonts}
\usepackage{amsthm}
\usepackage{booktabs}
\usepackage{algorithmic}
\usepackage[ruled, lined]{algorithm2e}
\usepackage{array}
\usepackage[caption=false,font=normalsize,labelfont=sf,textfont=sf]{subfig}
\usepackage{textcomp}
\usepackage{stfloats}
\usepackage{url}
\usepackage{verbatim}
\usepackage{graphicx}
\usepackage{cite}
\usepackage{bbm}
\usepackage{cases}
\usepackage{mathtools}
\usepackage{multirow}
\usepackage{comment}
\usepackage{color}
\usepackage{cleveref}       
\usepackage{pifont}
\newcommand{\cmark}{\ding{51}}%
\newcommand{\xmark}{\ding{55}}%

\DeclareMathOperator{\sgn}{sgn}

\newcommand{\sdf}{\phi_{g}}
\newcommand{\con}{\delta_{g}}
\newcommand{\segu}{u}
\newcommand{\conu}{\delta_{u}}
\newcommand{\conf}{\delta_{f}}
\newcommand{\flow}{\boldsymbol{F}_{g}}

\newcommand{\loss}{\mathcal{L}}
\newcommand{\uspace}{\mathbbm{U}}

\newcommand{\gtsp}{\Omega_{g}}

\newcommand{\sigm}{\mathcal{S}}
\newcommand{\qspace}{C^1}

\newcommand{\F}{\boldsymbol{F}}
\newcommand{\R}{\mathbbm{R}}
\newcommand{\N}{\mathbbm{N}}
\newcommand{\Nei}{\mathcal{N}}
\newcommand{\gt}{\boldsymbol{g}}

\newtheorem{defi}{Definition}
\newtheorem{thm}{Theorem}
\newtheorem{lem}{Lemma}
\newtheorem{cor}{Corollary}

\title{Contour Flow Constraint:  Preserving Global Shape Similarity for Deep Learning based Image Segmentation}

\date{}

\newif\ifuniqueAffiliation

\ifuniqueAffiliation 
\author{Shengzhe Chen \\
	\And
	Zhaoxuan Dong \\
    \AND
    Jun Liu \\
}
\else
\usepackage{authblk}

\author{Shengzhe Chen, Zhaoxuan Dong, Jun Liu \thanks{\texttt{This manuscript has been accepted by IEEE Transactions on Image Processing. The authors were with the School of Mathematical Sciences, Beijing Normal University when the manuscript was submitted. Email: \texttt{schen415@asu.edu} (Shengzhe Chen), \texttt{jliu@bnu.edu.cn} (Jun Liu).}}%
}

\renewcommand{\headeright}{}

\hypersetup{
pdftitle={A template for the arxiv style},
pdfsubject={q-bio.NC, q-bio.QM},
pdfauthor={David S.~Hippocampus, Elias D.~Striatum},
pdfkeywords={First keyword, Second keyword, More},
}

\begin{document}
\maketitle

\begin{abstract}
For effective image segmentation, it is crucial to employ constraints informed by prior knowledge about the characteristics of the areas to be segmented to yield favorable segmentation outcomes.
However, the existing methods have primarily focused on priors of specific properties or shapes, lacking consideration of the general global shape similarity from a Contour Flow perspective. Furthermore, naturally integrating this contour flow prior image segmentation model into the activation functions of deep convolutional networks through mathematical methods is currently unexplored.
In this paper, we establish a concept of global shape similarity based on the premise that two shapes exhibit comparable contours. Furthermore, we mathematically derive a contour flow constraint that ensures the preservation of global shape similarity.
We propose two implementations to integrate the constraint with deep neural networks. Firstly, the constraint is converted to a shape loss, which can be seamlessly incorporated into the training phase for any learning-based segmentation framework. Secondly, we add the constraint into a variational segmentation model and derive its iterative schemes for solution. The scheme is then unrolled to get the architecture of the proposed CFSSnet. Validation experiments on diverse datasets are conducted on classic benchmark deep network segmentation models. The results indicate a great improvement in segmentation accuracy and shape similarity for the proposed shape loss, showcasing the general adaptability of the proposed loss term regardless of specific network architectures. CFSSnet shows robustness in segmenting noise-contaminated images, and inherent capability to preserve global shape similarity.

\end{abstract}


\section{Introduction}

Image segmentation is a fundamental problem in image processing and has wide applications in medical image analysis and other areas. Its primary objective is to accurately segment regions of interest (foreground) within an image so that it is close and similar to the ground truth.

\begin{figure}[htb]
\centering
\includegraphics[width=0.8\linewidth]{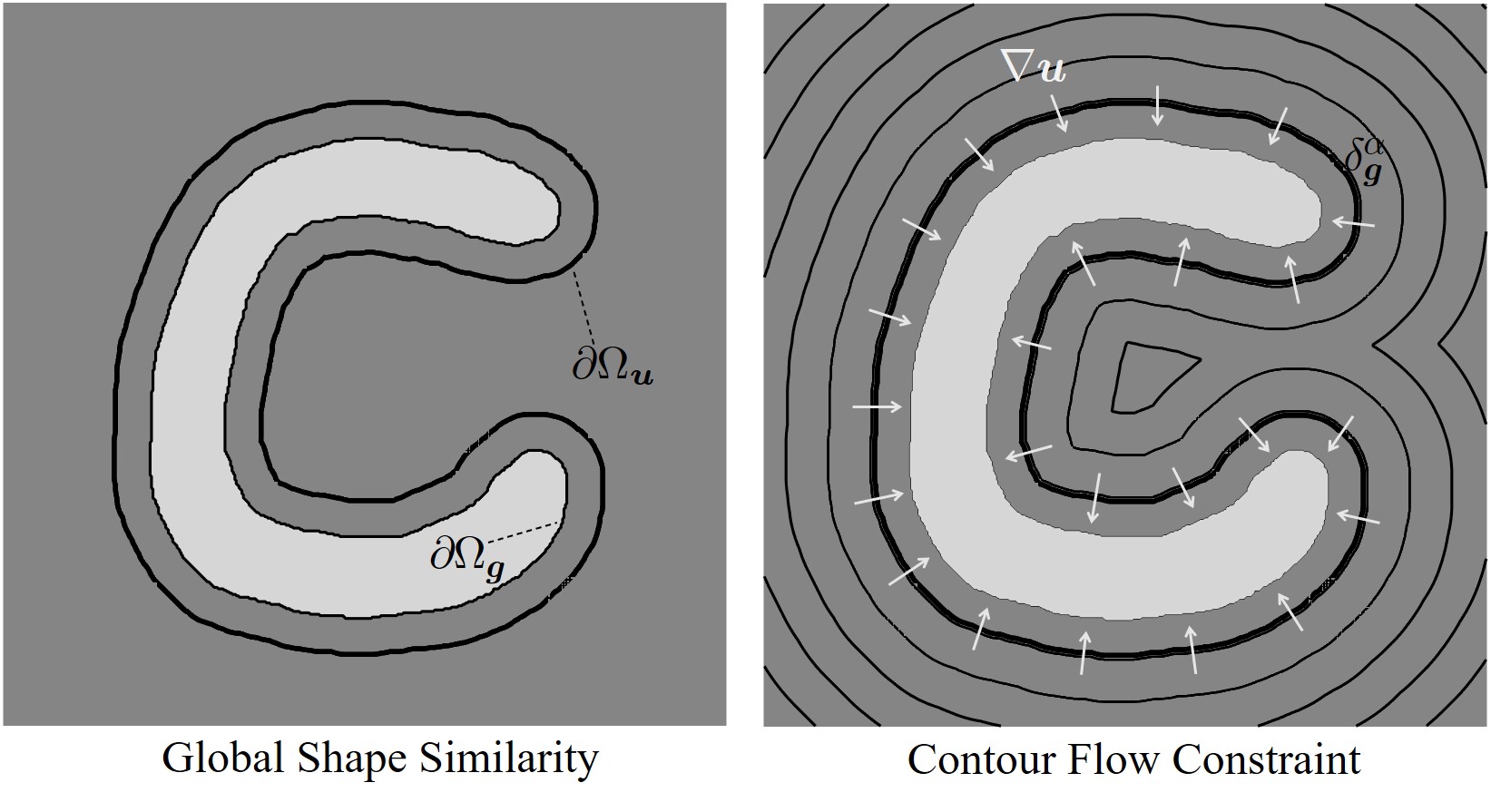}%
\caption{An intuition of the concept of global shape similarity of the ``C" letter.
}
\label{fig:intuition}
\end{figure}

In this paper, we mathematically derive the contour constraint condition that ensures the global shape similarity between the segmentation and the ground truth. Meanwhile, we propose two types of implementation to integrate the proposed shape constraint with deep neural networks, enabling them to maintain the shape characteristics and preserve global shape similarity. Specifically, we define a global shape similarity by two shapes sharing similar contours and equivalently convert such similarity into a constraint by the orthogonality of segmentation's gradient field and the ground truth contours' tangent field. This constraint is called the contour flow (CF) constraint to achieve global shape similarity. Please refer to Figure~\ref{fig:intuition} for an intuition of our idea. Then, we implement the CF constraint by two methods: (1) combine it into networks' training process to propose the regularized shape loss, which can be seamlessly integrated into any segmentation frameworks; (2) add it into a variational segmentation model to ensure global shape similarity property of its solution segmentation, then unroll the iterative schemes for the variational model to directly obtain deep network architectures, thereby proposed the unrolled CFSSnet, which can inherently capture shape characteristics within input image, and preserve global shape similarity.  

The main contributions and innovations of this paper can be concluded as:
\begin{itemize}
    \item We propose a contour flow constraint for image segmentation, which is mathematically derived to ensure global shape similarity in deep learning-based segmentation. This constraint provides a solid theoretical guarantee, offering a deeper understanding of how shape regularization improves segmentation performance.
    \item To fully utilize the prior knowledge of shape similarity within sample pairs during the training process, the proposed contour flow constraint is converted to the shape loss, which can be seamlessly integrated into any learning-based segmentation framework to enhance performance. Thereby, we propose a model-agnostic mechanism to preserve global shape similarity.
    \item To fully utilize the prior knowledge of shape similarity in prediction as hard constraints, we employ variational methods to incorporate shape constraints into the network architecture called CFSSnet, ensuring that the network outputs inherently preserve global shape similarity.
\end{itemize}

\section{Related work}

\subsection{Learning Based and Model Based Segmentation}

Traditionally, image segmentation can be tackled by learning-based methods and model-based methods.  Learning-based methods are data-driven. They utilize parameterized deep neural networks to learn the image features and classify foreground region pixels according to the ground truth. Fully Convolutional Network (FCN)~\citep{long2015fully}, U-net~\citep{ronneberger2015u}, DeepLab~\citep{chen2017deeplab}, and Vision-Transformer~\citep{dosovitskiy2020image} are popular learning-based segmentation frameworks with powerful learning and feature extraction ability. More recently, Segment Anything Model (SAM)~\citep{kirillov2023segment} with image encoder, prompt encoder, and mask decoder architecture has achieved awesome segmentation performances. In contrast, model-based methods, such as the Potts model~\citep{potts1952some}, Threshold Dynamics~\citep{merriman1994motion}, Active Contours~\citep{kass1988snakes}, Chan-Vese model~\citep{chan2001active}, employ variational models to minimize some energy associated with segmentation, so that the result segmentation encompasses a great fidelity. Importantly, specially designed regularization and prior-knowledge constraints could be added to the variational model to introduce desirable properties into the segmentation. Specifically, these priors include details of the area and shape of the segmented regions. \citep{veksler2008star} proposed a prior to constrain the segmentation region to be a star shape. \citep{royer2016convexity, yan2020convexity} both proposed a convexity shape constraint to segment convex regions based on Potts model~\citep{potts1952some} and Chan-Vese model~\citep{chan2001active} respectively. \citep{vicente2008graph} raised a prior to ensure topology connectivity of the segmentation region. ~\citep{rousson2002shape} utilized the level-set representation as a shape prior so that the segmented region exhibits global consistency with a prior shape. In recent years, there has been a trend towards hybrid approaches that combine the advantages of both learning-based and model-based methods, so that the segmentation framework could leverage the power of data-driven learning while incorporating prior knowledge or constraints to improve segmentation accuracy. In~\citep{liu2022deep}, the star shape prior was successfully integrated into deep convolution neural network (DCNN).  In the same paper, the authors also proposed a volume-preserving prior to constrain the area of the segmented region.  
The authors in~\citep{shit2021cldice} considered the topology of tabular, network-like regions and introduced a topology-preserving loss based on the structure prior.

\subsection{Shape Prior in Image Segmentation}
Shape prior plays a significant role in image segmentation, especially in medical imaging, where they help ensure the precise segmentation of tissues and organs. Many previous works have tried incorporating shape prior. 
In references \citep{boutillon2020combining, sadikine2024improving}, a convolutional auto-encoder is utilized to derive low-dimensional shape features, with an associated $L^2$ based shape loss applied within this reduced-dimensional space to address shape information. Nonetheless, the low-dimensional shape features yielded by the encoder are devoid of straightforward mathematical interpretability. Some works utilize contour information to assist segmentation, since contours are natural extensions of segmented region shapes, and can be utilized as shape prior.
In Dcan~\citep{chen2016dcan}, they emphasize accurate contour extraction by learning to segment boundaries alongside the main object, trying to encode shape information into the segmentation process. In Dmtn~\citep{tan2018deep}, they attempt to preserve shape by jointly learning task-specific and distance maps, a shape-related feature.
In Psi-net~\citep{murugesan2019psi} and Conv-MCD~\citep{murugesan2019conv}, they also use a similar joint-learning design that predicts boundary and distance map in multi-task decoders, aiming to improve segmentation accuracy by leveraging both forms of supervision. However, all these methods do not directly ensure contours of segmentation share similarity with ground truth shape, rather, contours are utilized as auxiliary information to assist the main segmentation process. More importantly, since contours information was implemented as some heuristically designed modules in DCNNs, they all lack strong mathematical guarantees on to what extent contours can help enhance segmentation.
Others also explored loss functions specifically designed to handle boundary adherence.
In~\citep{jurdi2021surprisingly}, the authors propose perimeter loss to employ a soft constraint that penalizes deviations from a target perimeter of an organ. In~\citep{wang2022active}, they propose active boundary loss to encourage the alignment between predicted and ground-truth boundaries through movements towards the predicted vector.
In TopNet~\citep{keshwani2020topnet}, to preserve the shape of the vessel centerlines in vessel segmentation, smoothing $L^1$
and topology-related loss functions were applied to a centerness score map. However, centerline similarity is only suitable for depicting elongated shapes such as vessels. 
Moreover, since there is no loss function as regularization during prediction, these methods cannot guarantee that the network maintains the regularized properties for individual images during the prediction process. 
Furthermore, using a loss function to impose constraints only addresses the overall characteristics of the sample set, without guaranteeing the preservation of shape features for each individual image.

To summarize shortcomings of previous works in terms of incorporating shape priors, most existing data-driven segmentation methods adopt an encoder to extract shape features from the output of the backbone network. They integrate shape priors by constraining the shape features extracted by the same encoder for ground truth segmentation through a regularization loss function. This approach has several drawbacks. First, the shape features extracted by the encoder are related to its training parameters and methods, and the extracted shape information lacks mathematical interpretability. Second, the regularization loss function can only constrain the entire sample set to roughly satisfy the shape priors, and it cannot guarantee that individual samples possess this shape feature. This is because the property of the encoder network to preserve shapes is unknown.

Therefore, we aim for the shape features extracted by the encoder to be mathematically interpretable, and for the regularization of these shape features to not only be effective roughly across the entire dataset but also to have an effect on each individual sample. This can be achieved by our proposed contour flow constraint with loss function and variational unrolling implementation.

\subsection{Unrolling}
The concept of unrolling, which has been explored in many works~\citep{ranftl2014deep,kobler2017variational,liu2022deep,li2022dual}, can be concluded as converting a variational model into DCNN, by expanding each iteration step for solution of the variational model into a layer in DCNN. The idea behind this technology is, 
for long, the design of deep network structures
lacks interpretability. Most networks are established based on researchers' experience without theoretical support. In contrast, variational models are built on solid mathematical facts and theoretical derivations, possessing good interpretability. Therefore,  with unrolling technology, the design of network structures is based on an appropriate mathematical model. In practice, once the iterative scheme to solve the variational model is obtained, one can replace certain operators with learnable kernels in DCNNs. More importantly, one can add constraint terms into a variation model like the problem (\ref{gvm}) to ensure desirable properties in the segmentation function. 
For example, in~\citep{liu2022deep}, Liu explored several terms to constrain overall shape properties like volume, piece-wise constant, and specific shape. These properties are inherited into the unrolled DCNNs.  

\section{The Proposed Global Shape Similarity Preserving Constraint}
\label{sec:CFC}

In this section, we derive the mathematical conditions for using the contour field as a measure of shape similarity.

\subsection{General Variational Segmentation Framework }
An image can be represented as a mapping $I: \Omega\mapsto\mathbbm{R}^d$, where $d$ is the image channels. 
Image segmentation aims to find a segmentation function $u:\Omega\mapsto\{0,1\}$ that accurately divides $\Omega$ into foreground $\{i\in\Omega | u(i)=1 \}$ and background $\{i\in\Omega | u(i)=0\}$ regions. 
We will utilize the variational segmentation framework~\citep{liu2022deep}, which bridges deep neural networks and variational optimization. As proposed in~\citep{liu2022deep}, we call this variational model as the soft threshold framework:
\begin{equation}
\label{gvm}
    u^* = \mathop{\arg\min}_{u\in\uspace} \left\{ \langle -o,u \rangle + \varepsilon\mathcal{H}(u)\right\},
\end{equation}
where $\uspace\coloneqq\{u|u\in C^1(\Omega), u: \Omega\mapsto[0,1]\}$ is the feasible domain for segmentation function. $o:\Omega\mapsto\R$ is a segmentation feature function, where the higher $o(i)$ indicates that the pixel $i$ is more likely to belong to the foreground. The inner product is defined as:
\begin{equation}
    \langle -o,u \rangle = \int_{\Omega} - o(i)u(i) di.
\end{equation}

Here the entropy term $\mathcal{H}(u) \coloneqq  \langle u, \ln{u} \rangle + \langle 1-u, \ln{(1-u)} \rangle$ and the parameter $\varepsilon>0$. The problem (\ref{gvm}) is strongly convex, and has a closed-form solution of the sigmoid function:
\begin{equation}
    u^* = \sigm(\frac{o}{\varepsilon})\coloneqq (1+\exp{(-\frac{o}{\varepsilon})})^{-1}.
\end{equation}

Since DCNNs also apply the sigmoid function to transform the last-layer output into probability, the problem (\ref{gvm}) can be perfectly combined with DCNNs by taking their last-layer output as $o$. 

\subsection{The Signed Distance Function for Ground Truth}
 
Let $\Omega_{g}\coloneqq\{ i\in\Omega | g(i) = 1 \}$ 
stand for ground truth, it can be equivalently defined by a signed distance function $\sdf$ derived from its boundary
$\partial\Omega_{g} \coloneqq \{ i\in\Omega_{g} | \forall \Nei(i), \exists j\notin\Omega_{g} \}$,
where $\Nei(i)$ stands for the adjacent pixels in $i$'s neighborhood.
\vspace{0.5em}
\begin{defi} 
\label{bdsim}
(Signed distance function \citep{chan2001active}) $\sdf$ measures signed distance between pixel $i$ to $\partial\Omega_{g}$. Specifically,   $\left|\sdf(i)\right| = D(i,\partial\Omega_{g})\coloneqq \underset{k\in\partial\Omega_{g}}{\min} \Vert i - k \Vert_2$ is (Euclidean) distance from pixel $i$ to boundary $\partial\Omega_{g}$. Its sign is determined by
\begin{equation}
	\label{sdf}
        \sgn{(\sdf(i))} =
	\left\{
	\begin{array}{ll}
		0, & i\in\partial\Omega_{g}, \\
		1, & i\in\Omega_{g}, \\
		-1, &  i\in\Omega\backslash\Omega_{g}.
	\end{array}
	\right.
\end{equation}
\end{defi}
\vspace{0.5em}
It is obvious $\Omega_{g} = \{i\in\Omega | \sdf(i)\ge 0 \}$. Generally, we assume $\sdf\in C^1(\Omega)$, which is regarded as the continuous form segmentation function of $\Omega_{g}$.

\subsection{Global Shape Similarity by Contours}

\begin{figure}
\centering
\includegraphics[width=1\linewidth]{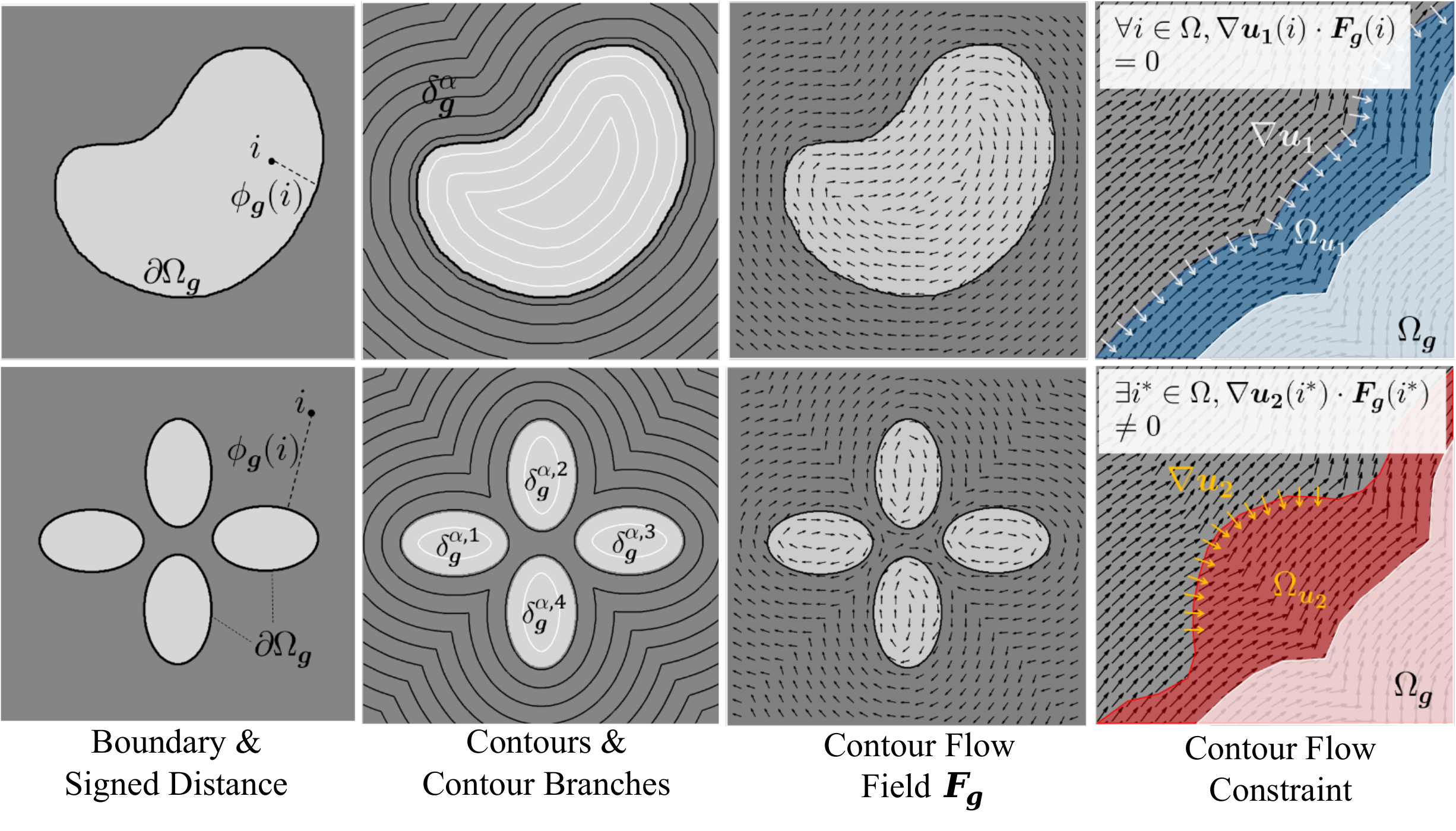}%
\caption{Intuitive illustration of some important concepts in this paper. 
}
\label{fig:contour}
\end{figure}

To ensure $u$ has similar shapes with $g$, we observe contours of $g$ naturally extend the shape $\Omega_{g}$ and exhibit similarity, as illustrated in Figure~\ref{fig:contour}. In other words, we are going to find $u$ which has similar contours with $g$. For example, two circles with the same center point and different radius have similar contours, and they are also similar to each other. 

Mathematically, $\alpha$-contour of $f: \Omega\mapsto\R,f\in C^1(\Omega)$ is defined as $\conf^{\alpha} = \{i\in\Omega|f(i)=\alpha\in\R\}$. However, $\conf^{\alpha}$ may be not a Jordan curve (simple closed curve) since it may consist of several connected components. For example, $\partial\Omega_{g}$, i.e. boundary of $\Omega_{g}$, is also the $0$-contour of $g$. It has more than one branch when $\Omega_{g}$ is a complex connected region (see Figure~\ref{fig:contour}). Therefore, we have the following definition: 
\vspace{0.5em}
\begin{defi} (Contour branch) $\forall \alpha\in\R$, the $\alpha$-contour of $f: \Omega\mapsto\R,f\in C^1(\Omega)$ is defined as $\conf^{\alpha} \coloneqq \{ i\in\Omega |f(i)=\alpha  \}$.  Further,
assume $\conf^{\alpha}$ has $N\in\N$  connected components, i.e. $\conf^{\alpha} = \bigcup_{n=1}^{N}\conf^{\alpha,n}$, disjoint union of $N$ branches. In case $\conf^{\alpha} = \emptyset$, let $N=1, \conf^{\alpha,n}= \emptyset$. Then, $\conf^{\alpha,n}$ is the $n$-th $\alpha$-contour branch of $f$.
\end{defi}
\vspace{0.5em}

By this definition, contour branches of $u,g\in C^1(\Omega)$ are $C^1$ Jordan curves. We denote $\conu^{\beta, m}, \con^{\alpha,n}$ as contour branches of $u,g$ respectively, where $\alpha,\beta\in\R, n\le N, m\le M$. To clarify, $\con^{\alpha,n}$ is actually derived from signed distance function $\sdf$, which means $\sdf(i) \equiv \alpha, i\in\con^{\alpha,n}$. Since contours are extensions of boundaries, sharing similar shape characteristics, if $u$ is going to be similar to $g$, they have to exhibit similarity within their contour branches. Therefore, we can define the so-called \textbf{contour similarity} for the two shapes as follows.

\vspace{0.5em}
\begin{defi} (Contour similarity)
\label{consim}
If $\forall \conu^{\beta,m}$, i.e. arbitrary contour branch of $u$, there exists corresponding contour branch $\con^{\alpha,n}$ of $g$, such that $\conu^{\beta,m} = \con^{\alpha,n}$, then $u$ has contour similarity with $g$, denoted as $\conu\sim \con$. This is also called the \textbf{global shape similarity} between $u,g$.
\end{defi}
\vspace{0.5em}

If $u$ has contour similarity with $g$, i.e. $\conu\sim \con$, it follows by an immediate corollary that the two also exhibit a \textbf{boundary similarity}. The corollary is demonstrated as follows.
\vspace{0.5em}
\begin{cor} 
\label{csim}
(Boundary similarity) Contour similarity $\conu\sim \con$ leads to boundary similarity between $\Omega_u,\Omega_{g}$. Specifically, for each branch $\conu^{\gamma,m}$ of $\partial\Omega_u$,  $\forall i,j\in \conu^{\gamma,m},$ $D(i,\partial\Omega_{g}) = D(j,\partial\Omega_{g}).$  $\partial\Omega_u, \partial\Omega_{g}$ are boundaries of $\Omega_u,\Omega_{g}$ respectively.

\end{cor}

\vspace{0.5em}

The proof of this Corollary can be found in the 
supplementary material of this paper.
Satisfying boundary similarity indicates pixels from the same branch of the segmentation boundary $\partial\Omega_u$ shares the same distance to the ground truth boundary $\partial\Omega_{g}$. Thus, the contour similarity in Definition~\ref{consim} ensures the shape $\Omega_u$ exhibits strong similarity with the ground truth $\Omega_{g}$. Next, we are going to derive an equivalent condition to constraint contour similarity between $u,g$.

\subsection{Contour Flow Constraint}

Contour similarity of $u,g$ requires $u$ to have similar contours with $g$. Intuitively, because their contour branches $\conu^{\beta, m}, \con^{\alpha,n}$ are $C^1$ Jordan curves, we can constrain $u,g$ have the same tangent vectors w.r.t the contour curve at each pixel. To make this constraint more practical to implement, we introduce an important lemma, which bridges contours and gradients.

\begin{lem} 
\label{lemma}
(Orthogonality of gradient-contour) Suppose $f: \Omega\mapsto\R,f\in C^1(\Omega), \nabla f(i) = (\frac{\partial f}{\partial x}(i),\frac{\partial f}{\partial y}(i))\in\R^2$. $\forall i\in\Omega,$ denote $\alpha = f(i)$, then the gradient vector $\nabla f(i)$ is orthogonal with $\alpha$-contour $\conf^{\alpha}\coloneqq\{j\in\Omega|f(j)=\alpha\}$ (at $i\in\Omega$), i.e. $\nabla f(i) \perp \conf^{\alpha}$.

\begin{proof}
See the supplementary material.
\end{proof}
\end{lem}

For $g$, we convert its contours into the tangent field, which we call the \textbf{contour flow field}. By Lemma~\ref{lemma}, since contours of $g$ are derived from signed distance function $\sdf$, it holds $\nabla \sdf(i)\perp \con^{\alpha,n}$, given $\sdf(i) = \alpha$. Therefore, by rotating $\nabla\sdf(i)$ with $\pi/2$, we get tangent of $\con^{\alpha,n}$ at $i$. See Figure~\ref{fig:contour} for an intuitive visualization of the field. We formally define the tangent field as:
\vspace{0.5em}
\begin{defi} 
\label{Tangent}
(Contour flow field) $\flow: \Omega\mapsto\R^2$ is called the contour flow field of $\Omega_{g}$. $\flow(i)$ is the unit tangent vector of $\con^{\alpha, n}$ at $i$ formulated by $\displaystyle  \flow(i) = R(\pi/2)\cdot\frac{\nabla\sdf(i)}{\Vert \nabla\sdf(i) \Vert}$, where $R(\cdot)$ is the rotation matrix $R(\theta) = \left(\begin{array}{lr}
   \cos{\theta} & -\sin{\theta} \\
    \sin{\theta} & \cos{\theta}
\end{array}\right)$.
\end{defi}
\vspace{0.5em}

With the contour flow field $\flow$, it is possible to constrain contour similarity between $u,g$ by a field orthogonality way. If $u$ has similar contours with $g$, then they should have the same contour flow fields. According to Lemma~\ref{lemma}, it leads to the gradient vector $\nabla u(i)$ orthogonal with $\flow(i)$. Importantly, this field orthogonality proves to be an equivalent condition of contour similarity in Definition~\ref{csim}. Our main theorem is demonstrated as follows:

\begin{thm} 
\label{th1} (Contour flow constraint) Suppose $u,\sdf\in C^1(\Omega)$, where $u$ is the segmentation function, and $\sdf$ is the signed distance function of $g$ in Definition~\ref{sdf}. Then, $u$ has contour similarity with $g$, i.e. $\conu\sim \con$, if and only if $\forall i\in\Omega$, $ \langle\nabla u(i),\flow(i)\rangle= 0 $, where $\nabla u:\Omega\mapsto\R^2$ is the gradient of $u$, and $\flow$ is the contour flow field in Definition~\ref{Tangent}.

\begin{proof}
See the supplementary material. 
\end{proof}
\end{thm}

Theorem~\ref{th1} reveals the relationships between contour similarity and contour flow field orthogonality. The intuition behind this theorem is illustrated in Figure~\ref{fig:contour}, fourth column: The white area in the bottom right corner of the two images represents the ground truth segmentation shape $\Omega_{g}$, while the black arrows depict the contour flow field $\flow$. In the image above, the blue area represents a certain segmentation shape $\Omega_{u_1}$, with its gradient vectors $\nabla u_1$ (white arrows) consistently orthogonal to the contour flow field, thereby exhibiting contour similarity to the ground truth. On the other hand, in the image below, the red area represents another segmentation shape $\Omega_{u_2}$, but its gradient vectors $\nabla u_2$ (gold arrows) are not always orthogonal to the contour flow field. This shape lacks contour similarity to the ground truth. Importantly, this condition is completely quantified and easy to implement. We call the orthogonality condition:
\begin{equation}
\label{cfc}
    \forall i\in\Omega, \langle\nabla u(i),\flow(i)\rangle = 0
\end{equation}
as the Contour Flow (CF) constraint for shape-preserving image segmentation. We will explore how to implement it into the segmentation process in the next sections. Specifically, we will propose two ways: (1) convert it to a loss function to add into any segmentation frameworks, (2) unroll a variational problem for segmentation, whose solution satisfies contour flow constraint condition, to generate a deep neural network preserving global shape similarity.

\section{Implementation I: Shape Loss for any Deep-learning based Segmentation Framework}
\label{sec:loss}
Widely used segmentation loss functions are cross entropy loss and dice loss defined on $u:\Omega\mapsto[0,1],g:\Omega\mapsto\{0,1\}$:
\begin{align}
\label{celoss}
    \loss_{CE} &= -\sum_{i\in\Omega}g(i)\log{u(i)} + (1-g(i))\log{(1-u(i))},\\
    \loss_{Dice} &= 1 - \frac{2\sum_{i\in\Omega}u(i)g(i)}{\sum_{i\in\Omega} u(i) + \sum_{i\in\Omega} g(i)}. \label{diceloss}
\end{align}

However, as analyzed in many works, these losses alone do not produce satisfying segmentation. The segmentation shape could suffer the ``salt and pepper'' noise, which means holes in inner shapes, and isolated points outside the main region. They only introduce pixel-wise fidelity, while important global properties between $\Omega_u$ and $\Omega_{g}$ like shape information are not included. Therefore, we utilize the CF constraint (Eq.\ref{cfc}) to propose a shape loss $\loss_{S}$ of cosine distance:
\begin{equation}
\label{shapeloss}
    \loss_{S} \coloneqq \sum_{i\in\Omega}\frac{\left| \langle\nabla u(i),\flow(i)\rangle \right|}{\Vert\nabla u(i) \Vert \cdot \Vert\flow(i) \Vert} = \sum_{i\in\Omega}\frac{\left| \langle\nabla u(i),\flow(i)\rangle \right|}{\Vert\nabla u(i) \Vert}.
\end{equation}

When $\loss_{S}\rightarrow 0$, it enforces $\langle\nabla u(i),\flow(i)\rangle\rightarrow 0$ at every $i\in\Omega$, which means field similarity of $\Omega_u$ with $\Omega_{g}$. According to Theorem~\ref{th1}, the global shape similarity will simultaneously be achieved. Because we actually require the orthogonality between vectors $\nabla u(i)$ and $\flow(i)$, the terms $\Vert\nabla u(i) \Vert$ is added to the denominator to make sure when $\Vert\nabla u(i) \Vert \rightarrow 0$, the loss goal can still be correctly optimized. 

Consequently, we propose the segmentation loss function to preserve global shape similarity, and will validate its effectiveness to enhance any DCNN-based segmentation frameworks in the experiments section:
\begin{equation}
    \loss = \alpha\loss_{Base} + \beta\loss_{S}, \ \alpha,\beta >0, Base\in \{CE, Dice\} 
\end{equation}

\section{Implementation II: Deep Neural Network by Unrolling Variational Model}
\label{sec:CFSS}
In this section, we will add the  CF constraint into a general variational segmentation model, i.e. the soft threshold classification framework, so that the output of DCNN will guarantee global shape similarity. Then, we will unroll the iteration schemes for the solution to generate a shape-preserving deep neural network.

\subsection{Variational Segmentation Model with Global Shape Similarity}

To add the CF constraint into a variational problem, we first convert Eq.(\ref{cfc}) into a dual formulation.

Let 
\begin{equation}
\label{innerabs}
   \mathbb{C}=\{u: \langle\nabla u(i),\flow(i)\rangle = 0,\forall i\in\Omega\},
\end{equation}

Then we have the following theorem: 
\begin{thm}\label{thm2}
Suppose $u\in C^2(\Omega),\flow\in C^1(\Omega)$ and $\flow\cdot \vec{\boldsymbol{n}}=0$ on $\partial\Omega$, then
\begin{equation*}
\max_{q\in C^1(\Omega)} \left\{\int_{\Omega}  u~{\rm div}(q\flow)  di\right\}=\left\{
\begin{array}{cc}
    0, & u\in \mathbb{C},\\
    +\infty,&\text{else}.
    \end{array}
\right.
\end{equation*}
\begin{proof}
    
     See the supplementary material.
\end{proof}
\end{thm}

Theorem \ref{thm2} states that the CF constraints set $\mathbb{C}$ can be replaced by the related $\max$ optimization problem.

Therefore, by adding the CF condition into the STD framework, we get the variational segmentation model with global shape similarity:
\begin{equation*}
    \min_{u\in\uspace\bigcap\mathbb{C}} \mathcal{E}(u,q) \coloneqq  \{ \langle -o,u \rangle + 
    \varepsilon\mathcal{H}(u)\}.
\end{equation*}
According to the theorem \ref{thm2}, the CF constraint set $\mathbb{C}$ can be
convert to a $\max$ problem, and one can get the following saddle problem
\begin{equation}
\label{primal}
    \min_{u\in\uspace} \max_{q\in\qspace} \mathcal{E}(u,q) \coloneqq  \{ \langle -o,u \rangle + 
    \varepsilon\mathcal{H}(u) + \langle {\rm div}(q\flow), u \rangle \}.
\end{equation}

We solve Eq.(\ref{primal}) by alternating iterative method of the following sub-problems:
\begin{subnumcases}{}
u^{t+1} = \mathop{\arg\min}_{u\in\uspace}\mathcal{E}(u,q^t) \label{subu},
 \\
q^{t+1} = \mathop{\arg\max}_{q\in\qspace}\mathcal{E}(u^{t+1},q), \label{subq}
\end{subnumcases}
in which the $u$ sub-problem (Eq.\ref{subu}) is formulated by:
$$u^{t+1} =\mathop{\arg\min}_{u\in\uspace} \left\{ \langle -o + {\rm div}(q^t\flow),u \rangle + \varepsilon\mathcal{H}(u)\right\}.$$

According to property of  framework~\citep{liu2022deep}, it has closed form solution, $u^{t+1} = \sigm((o - {\rm div}(q^t\flow))/\varepsilon)$, where $\sigm(\cdot)$ is sigmoid function. The $q$ sub-problem (Eq.\ref{subq}) is:
$$q^{t+1} =\mathop{\arg\max}_{q\in\qspace} \left\{ \langle q, -\langle \nabla u^{t+1}, \flow\rangle\rangle \right\}.$$

We solve the $q$ sub-problem by the gradient ascend method:
$q^{t+1} =q^t - \tau( \nabla u^{t+1}\cdot \flow)$, where $\tau>0$ is a update rate. As analyzed above, we actually require orthogonality between $\nabla u, \flow$. The $q$ variable here determines the bias term to be reduced on feature $o$ according to the extent of ``how orthogonal $\nabla u, \flow$ are''. When orthogonality is met, $q$ reaches optimal and stops to update. Therefore, at early steps, larger $\tau$ helps to converge to optimal value of $q$ more quickly. However, when $\nabla u\cdot\flow$ gets small, large $\tau$ may cause $q$ to ``overshoot'' and swing around the optimal. Empirically, setting $\tau =10$ can get a trade-off between speed and accuracy.

Consequently, we have the following schemes to solve the variational segmentation model with global shape similarity (Eq.\ref{primal}), and the corresponding Algorithm~\ref{algo1}:
\begin{subnumcases}{}
u^{t+1} = \sigm(\frac{o - {\rm div}(q^t\flow)}{\varepsilon}), \label{au}
 \\
q^{t+1} = q^t - \tau(\displaystyle{\nabla u^{t+1}}  \cdot \flow). \label{qu}
\end{subnumcases}

\begin{algorithm}
\caption{Algorithm of global shape similarity}\label{algo1}
\KwIn{segmentation feature $o$, contour flow $\flow$, setting hyperparameters $T$, $\varepsilon>0$, $\tau>0$.}
\textbf{Initialization:} $q^0 = \boldsymbol{0},t=0$\;
\While{$t < T$}{$\displaystyle u^{t+1}=\sigm(\frac{o - {\rm div}(q^t\flow)}{\varepsilon})$\;
$q^{t+1} = q^t - \tau\displaystyle{(\nabla u^{t+1}} \cdot\flow)$\;
$t\leftarrow t+1$\;}
\KwOut{Segmentation result $u=u^{T}$}
\end{algorithm}

\begin{figure}[htb]
\centering
\includegraphics[width=0.7\linewidth]{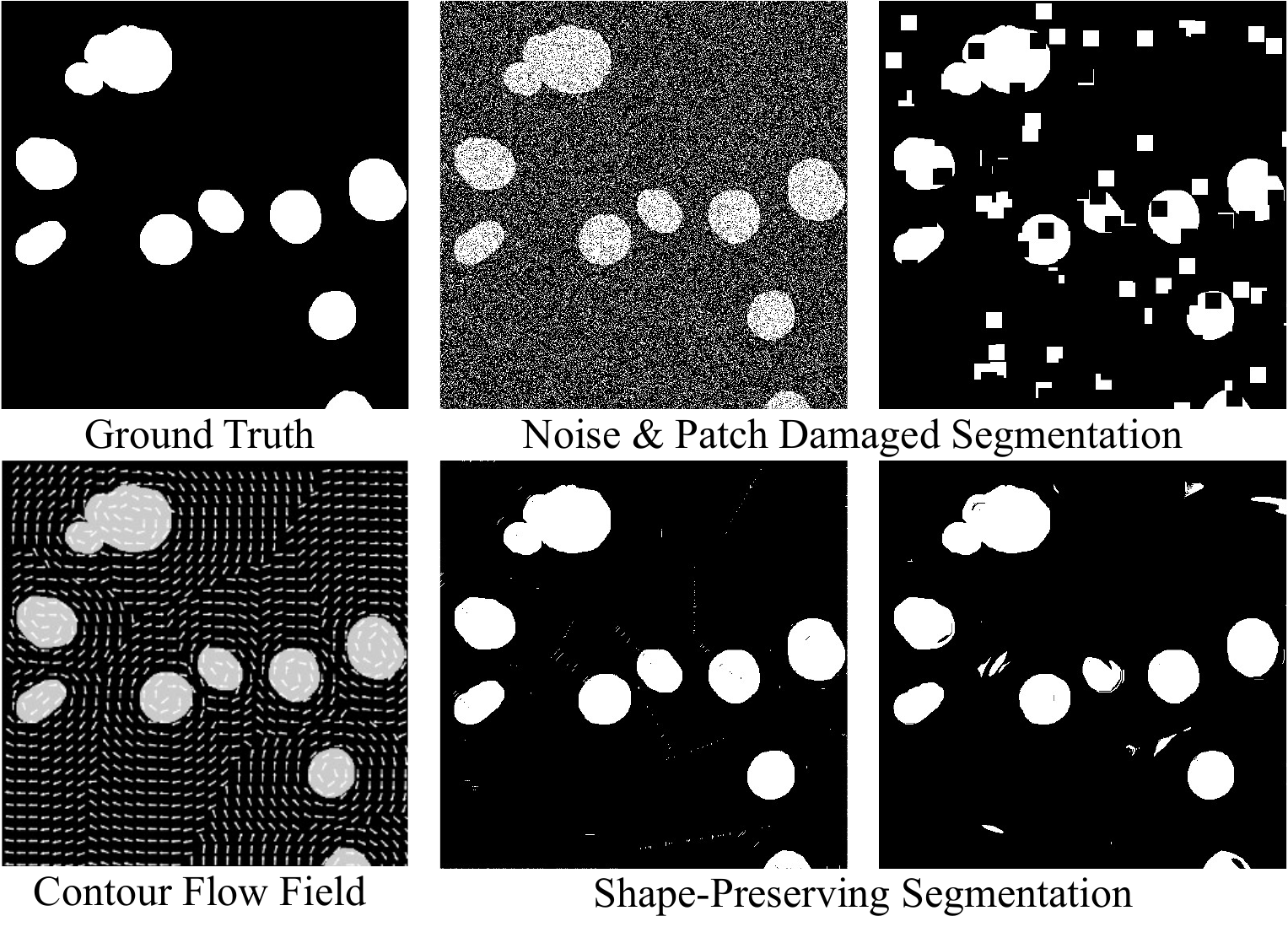}%
\caption{A toy example to illustrate the effectiveness of the proposed global shape similarity condition. 
}
\label{fig:toy}
\end{figure}

To validate the effectiveness of Algorithm~\ref{algo1}, we present a toy example in Figure~\ref{fig:toy}. A simple image $I$ with pixel value $0\&255$, i.e. black and white is constructed. The white region stands for the foreground, and the black region stands for the background. Its ground truth segmentation $g$ and contour flow field $\flow$ are shown in  Figure~\ref{fig:toy} in the first column. Then, $I$ is damaged in two ways: first, Gaussian noise $n$ is added to the image, i.e. $I_1 = I + n$. Second, random square patches $p$ with pixel values $0$ or $255$ are added, i.e. $I_2 = I + p$. For damaged images $I_1,I_2$, we use the K-means clustering to get their segmentation feature $o_1, o_2$, and the segmentation functions $u$ are given by $u_i=\sigm(o_i),i=1,2$. The segmentation results $u_1,u_2$ with threshold $u_i \ge 0.5$ are shown on first raw in Figure~\ref{fig:toy}. Then, we set $\varepsilon = \tau=10$, and input $o_1,o_2,\flow$ to Algorithm~\ref{algo1}. For the noise-damaged case, we set $T=100$ iterations, and for the patch-damaged case, $T=1000$ iterations. Finally, the output segmentation results with threshold $u^{T}_{i} \ge 0.5$ are shown on the second raw in Figure~\ref{fig:toy}. It can be seen that, in both cases, the image is damaged severely, and the segmentation results are horrible. However, given the contour flow $\flow$, by the proposed variational segmentation model with contour flow constraint, the output segmentation functions are recovered so that they exhibit global shape similarity with the ground truth.

It is important to note that in real segmentation scenarios, the ground truth $g$ and its contour flow field $\flow$ are unknown. However, by a deep learning way, we can get $\flow$ in the training phase. This inspires us to design a branch in the DCNN to predict a contour flow $\F$ by learning $\flow$, just like the way normal networks learn a segmentation function $u$ according to the ground truth $g$. Instead of post-processing predicted flow $\F$ with Algorithm~\ref{algo1}, we seamlessly integrate the algorithm into the network to make it end-to-end.  The architecture of our network does not come out of nowhere. It comes from unrolling the alternating iterative schemes (Algorithm~\ref{algo1}) for our variational model. Therefore, the network involves the shape-preserving mechanism by the architecture itself.

\subsection{CFSSnet -- Unrolled DCNN Preserving Global Shape Similarity}

\begin{figure*}
\centering
\includegraphics[width=1\textwidth]{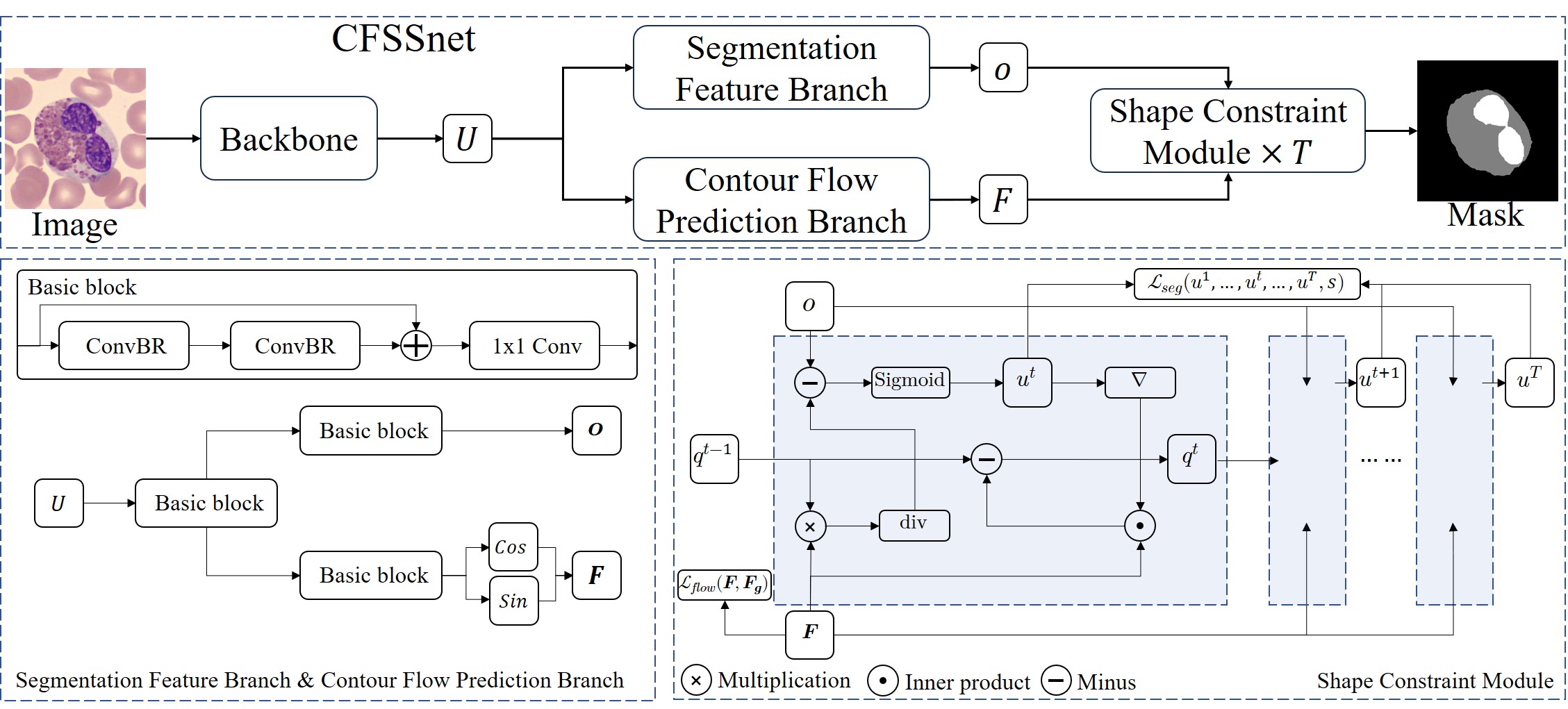}%
\caption{Architecture of CFSSnet and its components. ``ConvBR'' is short for ``Convolution-BatchNorm-ReLU''.
}
\label{fig:cfssnet}
\end{figure*}

We are now going to unroll the alternating iterative schemes for our proposed variational segmentation model with global shape similarity to obtain neural network architectures and propose the CFSSnet (Contour Flow Shape Similarity Networks) for the segmentation task. The whole architecture consists of three parts: backbone as feature extractor, two sub-branches to predict segmentation feature and contour flow and shape constraint modules to preserve shape similarity. Refer to Figure~\ref{fig:cfssnet} for the whole architecture and components.

\subsubsection{Backbone} The backbone extracts input image features for further processing. For implementation, we choose the U-structrue~\citep{ronneberger2015u} backbone, which proves to be very effective in integrating low and high levels features and is widely adopted for image segmentation tasks. Slightly different from baseline U-Net in Table~\ref{tab:result1}, we remove the batch-norm layer in up-sampling and the last $1\times 1$ convolution layer, so that it outputs $64$ channels feature map $U$.

\subsubsection{Segmentation Feature Branch \& Contour Flow Prediction Branch} 
Regarding the alternating iterative schemes (Eq.\ref{au}\&\ref{qu}), we need the segmentation feature $o$ and contour flow field $\flow$. For $o$, we regard it as features extracted by DCNN layers. For $\flow$, since it is derived from ground truth segmentation $g$, it cannot be obtained beforehand in forward propagation. However, since contour flow is also derived from $g$, it is possible to learn it through back-propagation in training, like the way the model learns $o$. Therefore, the output feature map of backbone, $U$, will be then processed in parallel by two branches: the feature branch producing segmentation feature $o$, and the flow branch predicting contour flow $\F$.

\subsubsection{Shape Constraint Module}
\label{sec:scm}
The shape constraint module is obtained by unrolling the alternating iterative schemes. The $\nabla, {\rm div}$ operators are implemented by convolution layers with fixed kernels of finite differences stencils representing discrete gradient and divergence operators. We use first-order scheme $\partial_x u = u(i_x+1,i_y) - u(i_x,i_y)$, and $\partial_y u= u(i_x,i_y+1) - u(i_x,i_y)$. The sigmoid function $\sigm$ corresponds to the sigmoid activation layer. According to Algorithm~\ref{algo1}, we use $T$ shape constraint modules in total. Each module outputs an intermediate segmentation function $u^t$, which will be involved in computing $u^{t+1}$ of the next module.

\subsubsection{Loss Function} 
The proposed CFSSnet has two parts of loss functions. For the first part, our network predicts a contour flow $\F$ for the segmentation shape, resembling true contour flow $\flow$. We use a $L^2$ loss to guide this learning process: 
\begin{equation}
    \loss_{flow} = \Vert \F - \flow \Vert_2.
\end{equation}

Although learning the contour flow $\flow$ encodes higher‐order (derivative) information that may seem more challenging than predicting the segmentation map itself, its main function is to impose a \emph{neighborhood‐based regularization} on the soft‐classification $u$. In a standard softmax segmentation network, each pixel is classified independently using only its local features $o$. Such an approach ignores how adjacent pixels should cohere. By introducing the additional CF prediction branch and minimizing the flow‐consistency loss we force the predicted flow $\F$ to align with the true gradient $\flow$, encouraging $u$ to respect its neighbors. The subsequent updates in Eqs.(\ref{au}) and (\ref{qu}) use this learned flow to refine $u$ in a way that naturally incorporates pixel‐neighborhood information. Consequently, even if $\F$ is not perfectly accurate, its presence still improves segmentation performance compared to a baseline that classifies pixels independently. We also have an ablation study to quantitatively investigate potential influence of flow estimation errors.

The second part loss is between the segmentation function $u$ and ground truth $g$. Since the network only handles a limited number of shape constraint modules, we back-propagate gradients of losses between ground truth $g$ and all the intermediate segmentation functions $u^t$:
\begin{equation}
    \loss_{seg} = \frac{1}{T}\sum_{t=1}^{T}(\loss_{CE}(u^t) + \loss_{S}(u^t))
\end{equation}
where $\loss_{CE}$ and $\loss_{S}$ are the cross-entropy loss (Eq.\ref{celoss}) and shape loss (Eq.\ref{shapeloss}) respectively defined in Section~\ref{sec:loss}. Consequently, the loss function for our proposed CFSSnet is formulated as:
\begin{equation}   
\loss_{CFSS} = \alpha\loss_{flow} + \beta\loss_{seg}, \ \alpha,\beta > 0.
\end{equation}
We will evaluate the performance of the proposed CFSSnet in the experiments section.

\section{Experiments}

In this section, we validate the CF constraint's effectiveness in preserving global shape similarity by the two implementations in section~\ref{sec:loss},\ref{sec:CFSS}. The weights in loss functions are set to $\alpha = \beta = 1$.  For the shape loss implementation, we add it to benchmark segmentation frameworks with baseline losses and compare the segmentation performances with and without the proposed shape loss. We would like to check whether shape loss can enhance global shape similarity despite specific network architectures. For the unrolled CFSSnet, we validate it by comparing its performance with benchmark networks trained with shape loss. To do so, we examine whether the inherent shape constraint modules can achieve better performance than fixed weights trained by the shape loss. Comparison experiments with the most related methods are also conducted.

\subsection{Datasets and Preprocessing}

This paper primarily validates and assesses the proposed method on several medical image datasets, which pose challenges in feature extraction and segmentation due to small differences in pixel grayscale values and the high requirement for global shape similarity. To evaluate performance on datasets of varying sizes, we utilize datasets comprising between 100 and 2700 images, each with different resolutions, as well as two additional widely used medical datasets. To ensure all test networks process images uniformly, we also resize images to suitable sizes.

\begin{itemize}
    \item \textbf{White Blood Cell Dataset (WBC1 and WBC2)}~\citep{zheng2018fast}: This dataset consists of white blood cell (WBC) images captured using electric autofocus and optical microscopes. It includes two subsets: WBC1 (300 small-sized images) and WBC2 (100 large-sized images), for evaluating cell segmentation, aiming to separate cell nuclei from cytoplasm. Images from WBC1 are enlarged from $120\times120$ to $128\times128$, and WBC2 from $300\times300$ to $320\times320$.
    \item \textbf{Retinal Fundus Glaucoma Dataset (REFUGE)}~\citep{orlando2020refuge}: The REFUGE dataset was developed for the 2018 MICCAI competition to support glaucoma diagnosis via deep learning. It contains 1200 retinal fundus images split evenly into training, validation, and test sets, focusing on segmenting the optic cup and optic disc. All images are cropped to $512\times512$ from the original $2048\times2048$.
    \item \textbf{Skin Lesion Dataset (ISIC)}~\citep{codella2018skin,tschandl2018ham10000}: From the ISIC 2018 competition, this dataset includes 2694 dermoscopic skin images, with 2594 images for training and 100 for testing. The goal is to delineate melanoma lesion areas for skin cancer diagnosis. All images are padded to a square shape and resized to $512\times512$.
    \item \textbf{Breast Ultrasound Dataset (BUSI)}~\citep{al2020dataset}: This dataset comprises 780 breast ultrasound images, used to segment benign and malignant breast cancer lesions.
    \item \textbf{Gastrointestinal Endoscopy Dataset (Kvasir)}~\citep{jha2019kvasir}: This dataset is from the 26th international conference of MultiMedia Modeling (MMM 2020), containing 1001 endoscopy images, and is used for segmenting polyps from gastrointestinal images.
\end{itemize}

No data augmentation techniques are applied aside from image resizing. For the training-validation-test split, REFUGE is already pre-divided into training, validation, and test sets.
For ISIC, 100 images are randomly selected from the training set for validation.
Regarding WBC, for WBC1, 60 images each are selected for validation and testing, for WBC2, 20 images each are chosen for testing and validation. For BUSI and Kvasir, 70\% images are split for training, 15\% for validation, and 15\% for testing. Both WBC and REFUGE datasets involve segmenting two classes of objects (cell nuclei and cytoplasm for WBC; optic cup and optic disc for REFUGE). During training, we utilize prior knowledge that the cell nuclei must be inside the cytoplasm, and the optic cup inside the optic disc, to transform the tasks into segmenting the entire cell or fundus and its internal structures.

\subsection{Evaluation Metrics}

Following section~\ref{sec:CFC}, let $\Omega_u$ and $\Omega_{g}$ denote the segmentation shapes of the network output and the ground truth respectively. $\partial\Omega_u$ and $\partial\Omega_{g}$ denote their corresponding segmentation boundaries. This paper employs three metrics to evaluate the results, each reflecting different aspects of segmentation performance, including shape overlapping, boundary discrepancy, and shape similarity.

\subsubsection{Dice Similarity Coefficient (\textbf{Dice})} Let $|\Omega_u|$ represent the number of pixels contained in $\Omega_u$. The \textbf{\textit{Dice}} score between $\Omega_u$ and $\Omega_{g}$ is defined as:
\begin{equation}
    {\rm \textbf{\textit{Dice}}}(\Omega_u,\Omega_{g}) = \frac{2*|\Omega_u\cap\Omega_{g}|}{|\Omega_u|+|\Omega_{g}|} \times 100\%.
\end{equation}

\textbf{\textit{Dice}} indicates the extent of overlap between the segmentation shape $\Omega_u$ and the ground truth. A larger value suggests that the segmentation shape $\Omega_u$ is closer to the ground truth.

\subsubsection{Boundary Distance (\textbf{BD})}

Let $|\partial\Omega_u|$ represent the number of pixels contained in $\partial\Omega_u$. The Boundary Distance (\textbf{\textit{BD}}) is defined as:
\begin{equation}
    {\rm \textbf{\textit{BD}}}(\partial\Omega_u,\partial\Omega_{g}) = \sum_{i\in\partial\Omega_u} \frac{D(i,\partial\Omega_{g})}{|\partial\Omega_u|},
\end{equation}
where $D(i,\partial\Omega_{g})=\mathop{\min}_{j\in\partial\Omega_{g}} \Vert i-j \Vert_2$, i.e. the distance from $i$ to the ground truth segmentation boundary $\partial\Omega_{g}$. This metric reflects the average distance from the segmentation boundary to the ground truth boundary. A smaller distance indicates that the boundaries are closer, implying more accurate results.

\subsubsection{Boundary Distance Standard Deviation (\textbf{BDSD})}

To characterize the shape similarity between the segmentation boundary $\partial\Omega_u$ and the ground truth boundary $\partial\Omega_{g}$, the Boundary Distance Standard Deviation (\textbf{\textit{BDSD}}) is defined as:
\begin{equation}
    {\rm \textbf{\textit{BDSD}}}(\partial\Omega_u,\partial\Omega_{g}) =  \sqrt{\frac{ \sum\limits_{i\in\partial\Omega_u} (D(i,\partial\Omega_{g})-{\rm \textbf{\textit{BD}}}(\partial\Omega_u,\partial\Omega_{g}))^2}{|\partial\Omega_u|}}.
\end{equation}

This is the square root of variance of $D(i,\partial\Omega_{g})$. If the segmentation boundary is too distinct, even a shape with small {\rm \textbf{\textit{BD}}} could have large {\rm \textbf{\textit{BDSD}}}. A smaller value of this metric indicates that the distances from the pixels on $\partial\Omega_u$ to  $\partial\Omega_{g}$ exhibit more consistency, suggesting a higher level of shape similarity between the two. If {\rm \textbf{\textit{BDSD}}} is equal to $0$, $\Omega_u, \Omega_{g}$ has exactly the shape similarity defined in Definition~\ref{consim}.

\subsection{Validate Effectiveness of the Shape Loss}
\label{sec:validatesl}

\begin{table}
  \centering
  \caption{Test results of benchmark frameworks trained with baseline losses and with the proposed shape loss on four datasets. 
  }
  \resizebox{\linewidth}{!}{
  \begin{tabular}[c]{cc|ccc|ccc|ccc|ccc}
  \hline
    \hline
    \multirow{2}{*}{\textbf{Networks}} & \multirow{2}{*}{\textbf{Loss Function}} & \multicolumn{3}{c}{\textbf{WBC1}} & \multicolumn{3}{c}{\textbf{WBC2}} & \multicolumn{3}{c}{\textbf{REFUGE}} & \multicolumn{3}{c}{\textbf{ISIC}} \\
    \cline{3-14}
    & & \textbf{\textit{Dice}}{$\uparrow$}& \textit{\textbf{BD}}{$\downarrow$} &\textit{\textbf{BDSD}}{$\downarrow$} & \textit{\textbf{Dice}}{$\uparrow$} & \textit{\textbf{BD}}{$\downarrow$}&\textit{\textbf{BDSD}}{$\downarrow$} & \textit{\textbf{Dice}}{$\uparrow$} & \textit{\textbf{BD}}{$\downarrow$}& \textit{\textbf{BDSD}}{$\downarrow$}& \textit{\textbf{Dice}}{$\uparrow$} & \textit{\textbf{BD}}{$\downarrow$}& \textit{\textbf{BDSD}}{$\downarrow$}\\
    \hline
    \multirow{4}{*}{FCN} & $\loss_{CE}$ & 95.89 & 1.362 & 1.589 & 93.50 & 5.513 & 6.164 &49.05 & 66.67 & 30.04 & 86.72 & 13.90 & 12.51\\
     & $\loss_{CE} + \loss_{S}$ & \textbf{96.00} & \textbf{0.924} & \textbf{0.843} & \textbf{94.22} & \textbf{4.999} & \textbf{4.770} & \textbf{50.81} & 67.74 & 33.52 & \textbf{87.01} & \textbf{12.07} & \textbf{10.08} \\
     \cline{2-14}
     & {$\loss_{Dice}$} & {95.56} & {1.317} & {1.438} & {94.71} & {5.186} & {5.907} & {65.79} & {39.92} & {22.73} & {84.80} & {19.09} & {16.82}\\
     & {$\loss_{Dice} + \loss_{S}$} & {95.55} & {\textbf{1.308}} & {\textbf{1.317}} & {\textbf{95.38}} & {\textbf{4.379}} & {\textbf{5.194}} & {\textbf{70.22}} & {\textbf{37.06}} & {25.50} & {\textbf{85.59}} & {\textbf{15.44}} & {\textbf{14.29}}\\
     \hline
     \multirow{4}{*}{UNET} & $\loss_{CE}$ & 96.22 & 1.392 & 1.678 & 95.65 & 4.185 & 5.493 & 57.56 & 68.54 & 46.48 & 86.80 & 18.35 & 19.26 \\
     & $\loss_{CE} + \loss_{S}$ & \textbf{96.73} & \textbf{0.840} & \textbf{0.896} & \textbf{95.91} & \textbf{3.308} & \textbf{3.357} & \textbf{64.74} & \textbf{46.92} & \textbf{35.94} & \textbf{87.06} & \textbf{15.02} & \textbf{15.45} \\
     \cline{2-14}
     & {$\loss_{Dice}$} & {96.50} & {1.167} & {1.580} & {94.58} & {5.579} & {6.296} & {65.06} & {47.51} & {30.66} & {83.99} & {22.21} & {20.62} \\
     & {$\loss_{Dice} + \loss_{S}$} & {\textbf{96.71}} & {\textbf{0.915}} & {\textbf{1.064}} & {\textbf{94.69}} & {\textbf{4.911}} & {\textbf{4.489}} & {\textbf{70.27}} & {\textbf{41.46}} & {\textbf{30.56}} & {\textbf{84.44}} & {\textbf{20.98}} & {\textbf{20.07}} \\
     \hline
     \multirow{4}{*}{GCN} & $\loss_{CE}$ & 94.19 & 1.629 & 1.611 & 93.30 & 6.148 & 6.965 & 50.39 & 71.91 & 29.09 & 87.29 & 13.09 & 11.78 \\
     & $\loss_{CE} + \loss_{S}$ & \textbf{95.13} & \textbf{1.168} & \textbf{1.019} & \textbf{94.52} & \textbf{4.033} & \textbf{4.151} & \textbf{66.16} & \textbf{39.42} & \textbf{16.10} & 87.09 & \textbf{12.00} & \textbf{9.249} \\
     \cline{2-14}
     & {$\loss_{Dice}$} & {95.61} & {1.663} & {2.136} & {95.52} & {5.164} & {7.190} & {69.67} & {36.37} & {17.91} & {86.31} & {13.63} & {11.00} \\
     & {$\loss_{Dice} + \loss_{S}$} & {\textbf{95.78}} & {\textbf{1.407}} & {\textbf{1.622}} & {94.02} & {\textbf{4.379}} & {\textbf{5.096}} & {\textbf{73.87}} & {\textbf{27.90}} & {\textbf{14.88}} & {\textbf{86.84}} & {\textbf{12.85}} & {11.10}\\
     \hline
     \multirow{4}{*}{DeepLabv3p} & $\loss_{CE}$ & 95.04 & 1.498 & 1.680 & 95.17 & 5.146 & 6.729 & 76.92 & 20.42 & 12.03 & 87.91 & 11.72 & 9.824\\
     & $\loss_{CE} + \loss_{S}$ & \textbf{95.75} & \textbf{1.096} & \textbf{1.113} & \textbf{96.14} & \textbf{3.121} & \textbf{3.849} & \textbf{80.50} & \textbf{16.52} & \textbf{7.854} & \textbf{88.22} & \textbf{10.77} & \textbf{9.540} \\
     \cline{2-14}
     & {$\loss_{Dice}$} & {94.97} & {1.741} & {2.141} & {94.99} & {5.841} & {7.624} & {81.02} & {17.77} & {9.791} & {87.43} & {11.74} & {9.734} \\
     & {$\loss_{Dice} + \loss_{S}$} & {\textbf{95.84}} & {\textbf{1.252}} & {\textbf{1.478}} & {\textbf{95.69}} & {\textbf{4.368}} & {\textbf{5.493}} & {\textbf{84.33}} & {\textbf{13.00}} & {\textbf{8.079}} & {\textbf{88.21}} & {\textbf{10.78}} & {\textbf{9.306}}\\
     \hline
     \multirow{4}{*}{UNET++} & $\loss_{CE}$ & 96.71 & 1.030 & 1.225 & 93.94 & 6.896 & 8.148 & 65.82 & 47.56 & 32.04 & 85.32 & 16.59 & 16.38 \\
     & $\loss_{CE} + \loss_{S}$ & 96.46 & \textbf{1.020} & \textbf{1.137} & \textbf{95.54} & \textbf{3.550} & \textbf{4.105} & \textbf{66.90} & \textbf{44.32} & 33.70 & \textbf{85.74} & \textbf{15.82} & \textbf{15.78} \\
     \cline{2-14}
     & {$\loss_{Dice}$} & {96.54} & {1.082} & {1.261} & {94.36} & {6.144} & {7.340} & {66.07} & {52.66} & {39.12} & {84.76} & {22.67} & {22.34} \\
     & {$\loss_{Dice} + \loss_{S}$} & {\textbf{96.72}} & {\textbf{1.016}} & {\textbf{1.220}} & {\textbf{95.50}} & {\textbf{3.647}} & {\textbf{4.131}} & {\textbf{71.59}} & {\textbf{40.15}} & {\textbf{29.47}} & {84.47} & {\textbf{20.69}} & {\textbf{20.31}}\\
     \hline
     \multirow{4}{*}{HRNET} & $\loss_{CE}$ & 94.22 & 1.601 & 1.593 & 93.12 & 6.242 & 6.807 & 65.19 & 85.80 & 48.14 & 83.95 & 19.49 & 18.23 \\
     & $\loss_{CE} + \loss_{S}$ & \textbf{94.57} & \textbf{1.462} & \textbf{1.398} & \textbf{94.95} & \textbf{4.345} & \textbf{4.751} & \textbf{66.65} & \textbf{39.49} & \textbf{17.69} & \textbf{85.88} & \textbf{15.68} & \textbf{13.86} \\
     \cline{2-14}
     & {$\loss_{Dice}$} & {95.14} & {1.831} & {2.199} & {94.31} & {5.685} & {7.432} & {67.32} & {34.54} & {19.46} & {84.85} & {18.61} & {17.64} \\
     & {$\loss_{Dice} + \loss_{S}$} & {\textbf{95.56}} & {\textbf{1.384}} & {\textbf{1.476}} & {94.12} & {\textbf{4.834}} & {\textbf{4.695}} & {\textbf{68.60}} & {35.05} & {21.55} & {\textbf{85.06}} & {\textbf{15.78}} & {\textbf{14.29}} \\
     \hline
     \multirow{2}{*}{\textbf{CFSSnet}} & {$\loss_{CE} + \loss_{flow}$} & 96.21 & 1.264 & 1.576 & 96.54 & 2.967 & {4.172} & {82.54} & {14.69} & {8.833} & {88.00} & {14.52} & {10.60} \\
     & {$\loss_{seg} + \loss_{flow}$} & {\textbf{96.70}} & {\textbf{0.802}} & {\textbf{0.819}} & {95.68} & {3.118} & {\textbf{2.839}} & {\textbf{85.10}} & {\textbf{13.01}} & {\textbf{7.378}} & {\textbf{88.53}} & {\textbf{13.00}} & {\textbf{9.594}}\\
     \hline
     \hline
  \end{tabular}
  }
  \label{tab:result1}
\end{table}

To validate the effectiveness of the proposed shape loss in preserving global shape similarity between the network's segmentation and the ground truth, we design the following experiments. First, we train the benchmark segmentation frameworks with two baseline loss functions, i.e. cross-entropy loss $\loss_{CE}$ and dice loss $\loss_{Dice}$ (soft dice loss without squared denominator as defined in Eq.(\ref{diceloss})). Then, we train the networks once again with the same setups but adding the shape loss $\loss_{S}$ to baseline losses.
Finally, we evaluate and compare their performances on test datasets by the three metrics, namely \textbf{\textit{Dice}}, \textbf{\textit{BD}}, \textbf{\textit{BDSD}}. By the comparison, we shall see whether networks trained with the shape loss exhibit better global shape similarity and whether such an effect depends on specific network architecture.

\subsubsection{Benchmark Segmentation Frameworks} We select several classic segmentation frameworks from open-source implementations, including FCN~\citep{long2015fully}, U-net~\citep{ronneberger2015u}, GCN~\citep{peng2017large}, DeepLabv3plus~\citep{chen2018encoder}, UNET++~\citep{zhou2018unet++}, and HRNet~\citep{sun2019deep}. These frameworks are designed with distinct architectures.

\subsubsection{Training Setups}

The epochs of training are set to 100 epochs for the ISIC dataset and 200 epochs for all other datasets. The learning rates are set to $1e-3$ for WBC1, $1e-3$ for WBC2, $1e-4$ for REFUGE, and $1e-5$ for ISIC. The batch size for training on all datasets is set to 10.

For all the deep networks used in the experiments, we employ the default network layer parameter initialization method, which is the Kaiming initialization~\citep{he2015delving}. 
Subsequently, two types of training are conducted for each dataset, baseline network and baseline loss function: the first trained with baseline loss, and the second with baseline loss adding shape loss. Finally, segmentation shapes are generated on the test sets. We then compute evaluation metrics and perform comparisons and analyses for the two types of training.

\subsubsection{Performances and Analyses}

\begin{figure*}[!htb]
\centering
\includegraphics[width=1\textwidth]{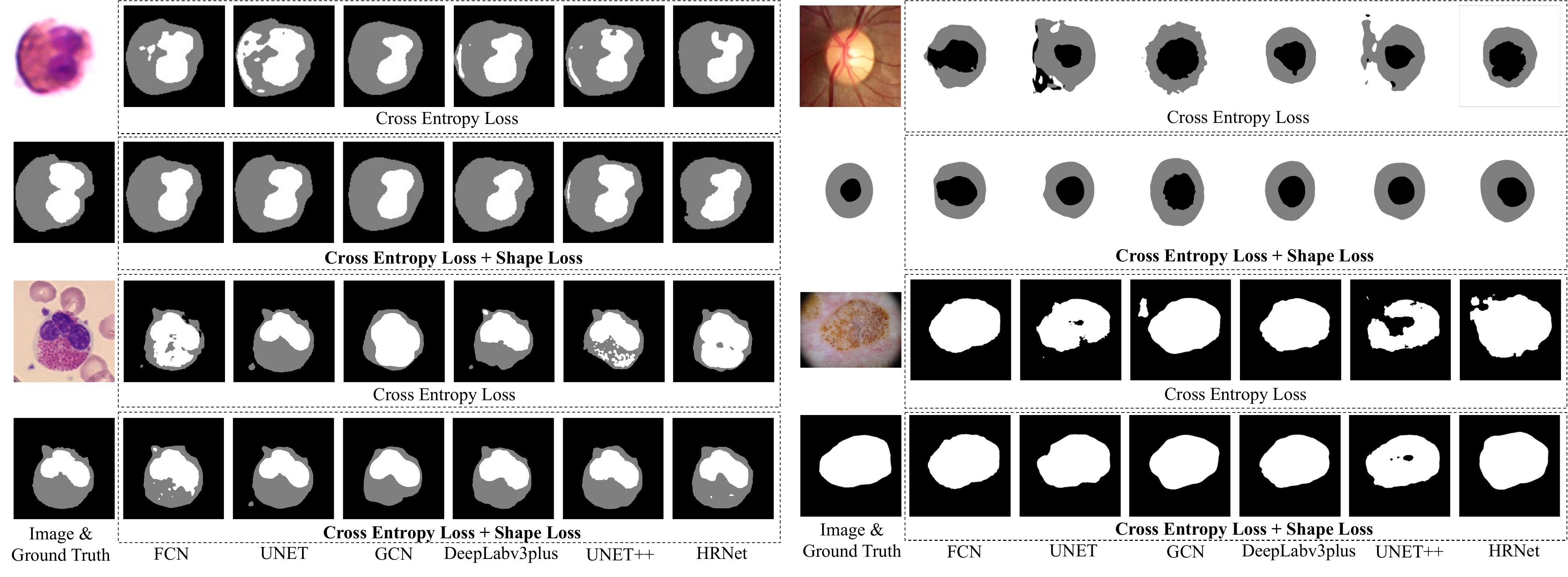}%
\caption{
Segmentation results of benchmark networks trained with cross entropy loss alone and together with the proposed shape loss.
}
\label{fig:shapelossresult}
\end{figure*}

Table~\ref{tab:result1} presents the segmentation performance results of the benchmark networks on the test sets of various datasets. With incorporation of the proposed shape loss, the network's performance significantly improves on all datasets with few exceptions. Figures~\ref{fig:cfssresult} provides a more intuitive visualization, showing that on the same test image, the networks trained with the shape loss produce segmentation results that are much closer to the ground truth in terms of shape. Moreover, these results exhibit fewer salt-and-pepper noise, isolated points, and erroneous regions. This indicates that training with the shape loss allows the network parameters to capture the global shape information of objects, resulting in improved segmentation performance. These comparative experiments also demonstrate that the proposed shape loss is not dependent on specific network architectures or baseline loss functions when applied in practice. As long as this loss term is incorporated into the network training process, segmentation performance can be improved. In essence, it can be seamlessly integrated into arbitrary networks' training progress, enhancing segmentation performance across the board, which showcases the proposed shape loss's generalizability.

\subsection{Validate Inherent Shape Preserving Mechanism of the Unrolled CFSSnet}
\label{sec:validatecfss}
The CFSSnet is obtained by unrolling the alternating iterative algorithm for the variational segmentation model with global shape similarity. In the architecture, several shape constraint modules are applied to refine the segmentation function. Therefore, we conduct experiments to examine whether the shape constraint modules can inherently achieve global shape similarity according to the input image.

\subsubsection{Testsets}

To examine the inherent shape-preserving ability of the CFSSnet unrolled from the variational model, and to compare with those directly incorporating the shape loss, we include perturbed test sets for the datasets used. Specifically, Gaussian noise and salt-and-pepper noise are added to the original WBC1 and WBC2 datasets to create test sets with perturbance. Gaussian noise follows a Gaussian distribution with a mean of 0 and a variance of 20, while salt-and-pepper noise introduces white (255) and black (0) pixels to 2\% of the image pixels. This results in four test sets denoted as WBC1-GS, WBC1-SP, WBC2-GS, and WBC2-SP. The test sets with "GS" suffixes have added Gaussian noise, while those with "SP" have added salt-and-pepper noise.

\subsubsection{Training Setups}

The relaxation factor $\varepsilon$ in the shape constraint module is set to 10, the gradient update rate $\tau$ is set to 10, and a total of $T=20$ shape constraint modules are used. The training process employs the Adam optimizer with default parameters, and the learning rate is set to $1e-3$. Training is conducted for 200 epochs. After the training of CFSSnet is completed, comparative experiments are conducted with the DeepLabv3plus model trained with the shape loss on the four test sets to assess and compare the segmentation performance.

\subsubsection{Performances and Analyses}

\begin{figure*}[htb]
\centering
\includegraphics[width=1\textwidth]{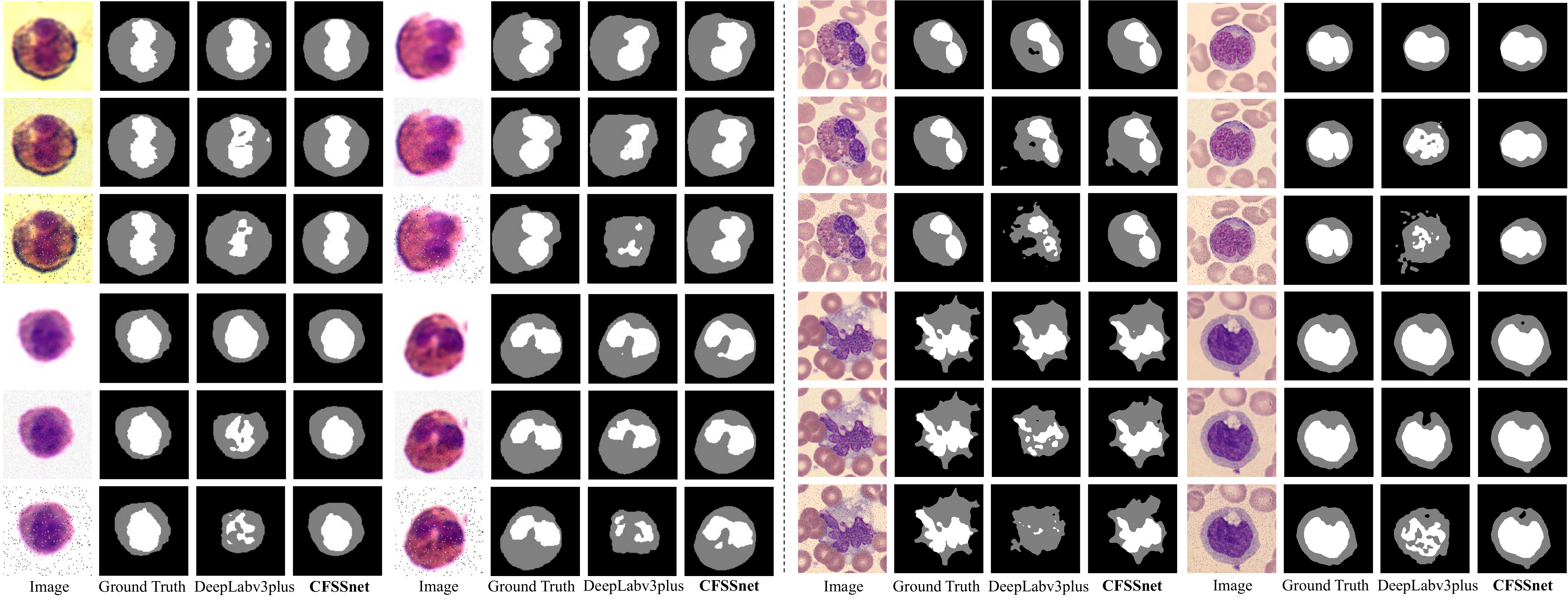}%
\caption{
Comparison of DeepLabv3plus with proposed shape loss and CFSSnet on perturbed WBC datasets.
}
\label{fig:cfssresult}
\end{figure*}

Table~\ref{tab:result2} presents the test results on various datasets. It can be observed that the proposed CFSSnet exhibits greater robustness to noise. This is because CFSSnet enforces global shape similarity constraints through its network architecture, which is derived from the iterative schemes of the variational model through unrolling. Consequently, during the forward propagation, the network essentially performs the iteration schemes, so that the output results inherently possess the properties of the shape similarity constraint added to the variational model. In contrast to training with shape loss, where the networks learn the correct shape information and perform well on the original test set, they struggle to adapt to the new image distribution with noise. Figure~\ref{fig:cfssresult} provides a visual comparison and demonstration of the experimental results.

\begin{table}[htb]
    \centering
  \caption{Test results of DeepLabv3plus \& CFSSnet on the perturbed WBC datasets. GS stands for Gaussian noise, and SP represents salt-and-pepper noise.} 
  \begin{tabular}[c]{c|ccc|ccc}
  \hline
    \hline
        \multirow{2}{*}{\textbf{Network}}  & \multicolumn{3}{c|}{\textbf{WBC1-GS (test-only)}} & \multicolumn{3}{c}{\textbf{WBC1-SP (test-only)}}\\ 
       & \textbf{\textit{Dice}}{$\uparrow$} & \textbf{\textit{BD}}{$\downarrow$} & \textbf{\textit{BDSD}}{$\downarrow$} & \textbf{\textit{Dice}}{$\uparrow$} & \textbf{\textit{BD}}{$\downarrow$} & \textbf{\textit{BDSD}}{$\downarrow$}  \\
       \hline
       DeepLabv3p & 94.12 & 1.466 & 1.396 & 84.23 & 3.415 & 2.622 \\
       \textbf{CFSSnet} & \textbf{96.46} & \textbf{0.826} & \textbf{0.802} & \textbf{96.12} & \textbf{0.899} & \textbf{0.871} \\
    \hline
    \hline
    \multirow{2}{*}{\textbf{Network}}  & \multicolumn{3}{c|}{\textbf{WBC2-GS (test-only)}} & \multicolumn{3}{c}{\textbf{WBC2-SP (test-only)}}\\
    & \textbf{\textit{Dice}}{$\uparrow$} & \textbf{\textit{BD}}{$\downarrow$} & \textbf{\textit{BDSD}}{$\downarrow$} & \textbf{\textit{Dice}}{$\uparrow$} & \textbf{\textit{BD}}{$\downarrow$} & \textbf{\textit{BDSD}}{$\downarrow$}  \\
    \hline
    DeepLabv3p & 90.48 & 7.130 & 6.696 & 81.23 & 11.05 & 9.827 \\
    \textbf{CFSSnet} & \textbf{95.68} & \textbf{3.156} & \textbf{3.652} & \textbf{95.90} & \textbf{2.810} & \textbf{3.151} \\
    \hline
    \hline
    \end{tabular}
  \label{tab:result2}
\end{table}

\subsection{3D Segmentation with the Proposed Method.}

\begin{figure}
\centering
\includegraphics[width=0.7\linewidth]{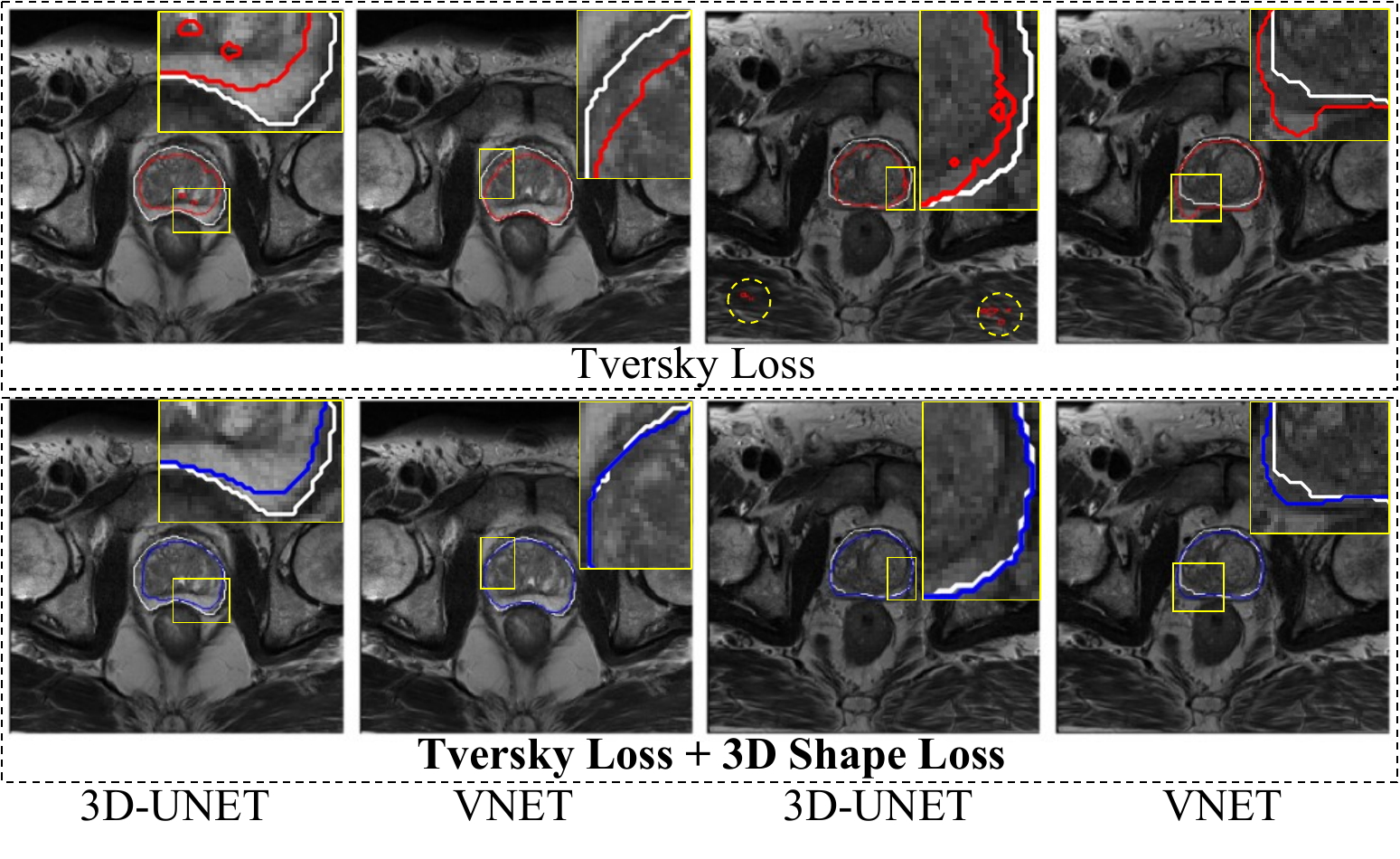}%
\caption{Comparison between 3D segmentation (ground truth colored in white) on prostate MR images trained with Tversky loss alone (colored in red) and together with 3D shape loss (colored in blue).}
\label{fig:3Dresult}
\end{figure}

In the field of medical image processing, 3D segmentation is commonly used to handle the detailed structures in medical images. We find that, by using an equivalent constraint $\nabla u \times \nabla \sdf = \boldsymbol{0}$, our proposal can also implemented in 3D. For some preliminary results, We carried out 3D segmentation experiments on PROMISE2012 challenge dataset~\citep{litjens2014evaluation} with two baseline networks, 3D-Unet~\citep{cciccek20163d} and V-net~\citep{milletari2016v}. Likewise in section~\ref{sec:validatesl}, for each network, three types of loss, namely the weighted CE loss $\loss_{WCE}$, dice loss $\loss_{Dice}$, and Tversky loss~\citep{salehi2017tversky} $\loss_{Tversky}$  are applied and then compared with adding our 3D shape loss $\loss_{3DS} = \Vert \nabla u \times \nabla \sdf \Vert $. The results are shown in Table~\ref{tab:result3D}.  Refer to Figure~\ref{fig:3Dresult} for intuitions of our method's advantages in maintaining boundary shapes and eliminating outer isolated regions or inner holes.

\begin{table}[htb]
    \centering
  \caption{Preliminary results of 3D Shape Loss on baseline models} 
  \begin{tabular}[c]{cc|ccc}
  \hline
    \hline
     \multirow{2}{*}{\textbf{Network}} & \multirow{2}{*}{\textbf{Loss Function}} & \multicolumn{3}{c}{\textbf{PROMISE2012}} \\
      & & \textbf{\textit{Dice}}{$\uparrow$} & \textbf{\textit{BD}}{$\downarrow$} & \textbf{\textit{BDSD}}{$\downarrow$} \\
    \hline
    \hline
     \multirow{6}{*}{3D-Unet} & $\loss_{WCE}$ & 66.49 & 6.120 & 4.940 \\
     & $\loss_{WCE} + \loss_{3DS}$ & \textbf{68.61} & \textbf{2.982} & \textbf{2.185} \\
     \cline{2-5}
      & $\loss_{Dice}$ & 74.59 & 8.897 & 11.25 \\
     & $\loss_{Dice} + \loss_{3DS}$ & \textbf{79.51} & \textbf{2.235} & \textbf{2.381} \\
      \cline{2-5}
      & $\loss_{Tversky}$ & 78.39 & 6.903 & 9.295 \\
     & $\loss_{Tversky} + \loss_{3DS}$ & \textbf{82.21} & \textbf{1.641} & \textbf{1.437} \\
     \cline{1-5}
      \multirow{6}{*}{V-Unet} & $\loss_{WCE}$ & 69.37 & 4.579 & 4.810 \\
     & $\loss_{WCE} + \loss_{3DS}$ & \textbf{77.39} & \textbf{2.211} & \textbf{1.736} \\
     \cline{2-5}
      & $\loss_{Dice}$ & 65.02 & 3.245 & 5.285 \\
     & $\loss_{Dice} + \loss_{3DS}$ & \textbf{80.82} & \textbf{1.744} & \textbf{1.501} \\
      \cline{2-5}
      & $\loss_{Tversky}$ & 77.68 & 3.938 & 4.852 \\
     & $\loss_{Tversky} + \loss_{3DS}$ & \textbf{81.53} & \textbf{1.785} & \textbf{1.731} \\
    \hline
    \hline
  \end{tabular}
  \label{tab:result3D}
\end{table}

\subsection{Comparison Experiments}

\begin{figure*}
\centering
\includegraphics[width=1\textwidth]{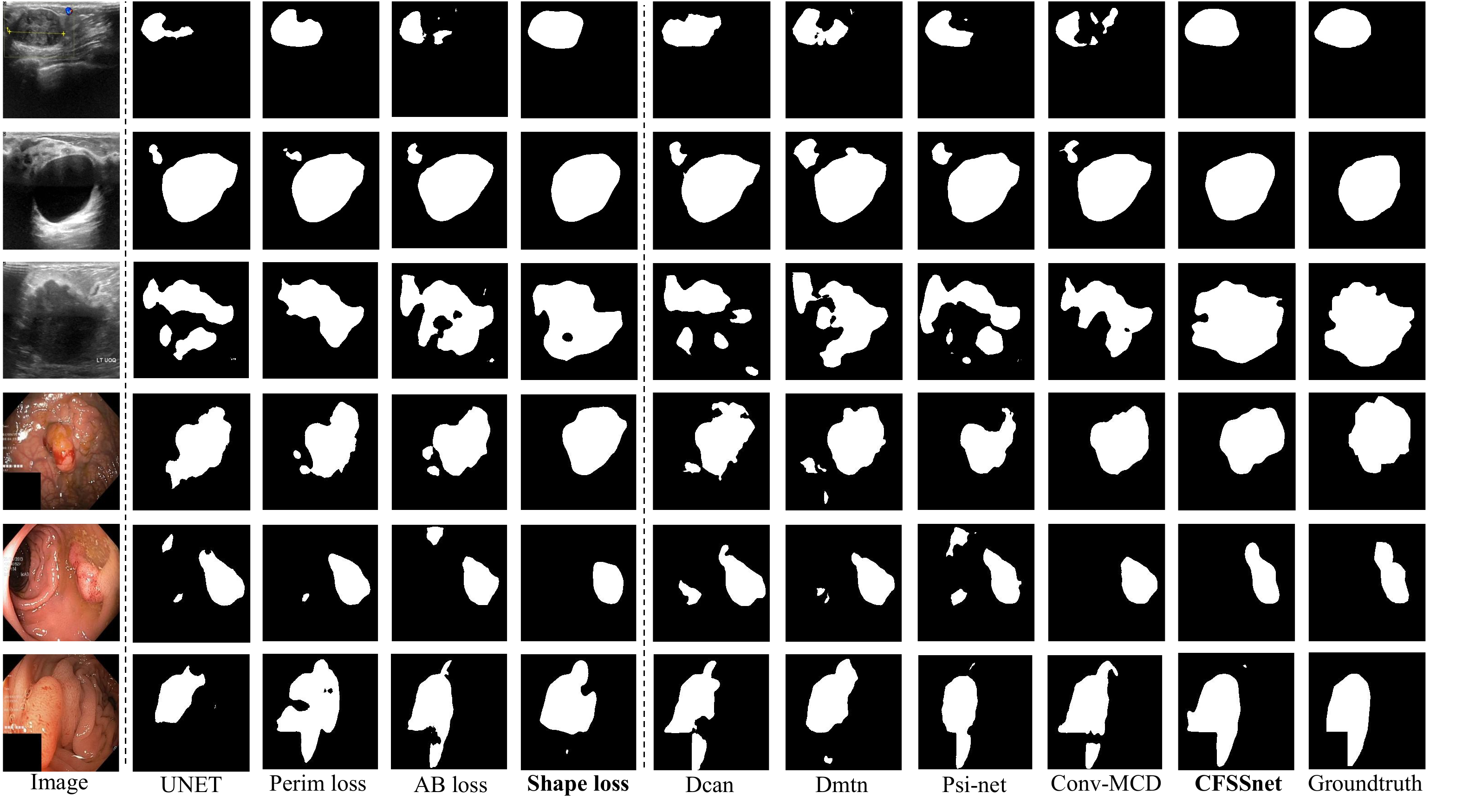}%
\caption{Comparison of performances with the most related methods. 
}
\label{fig:comparison}
\end{figure*}

In this section, we compare the segmentation performance of our proposed methods against several most related state-of-the-art techniques. On BUSI and Kvasir datasets.

\subsubsection{Reference Methods} We compare our methods, i.e. $\loss_{S}$ shape loss and CFSSnet with the most related SOTA methods, including Dcan~\citep{chen2016dcan}, Dmtn~\citep{tan2018deep}, Psi-net~\citep{murugesan2019psi}, Conv-MCD~\citep{murugesan2019conv}, $\loss_{perim}$ perimeter loss~\citep{jurdi2021surprisingly}, $\loss_{AB}$ active boundary loss~\citep{wang2022active}. These methods all considered incorporating contours, distance maps, or boundary information as shape prior for better segmentation, and have demonstrated competitive performance on similar image segmentation tasks. To ensure a fair comparison, we adopt the same training and testing protocols across all methods. U-net~\citep{ronneberger2015u} is chosen as a baseline model for methods of adding losses.  The models are initialized by trained with 150 epochs, batch sizes of 16, and a learning rate of $1e-4$. Then, all methods are evaluated on testset by the aforementioned metrics \textbf{\textit{Dice}}, \textbf{\textit{BD}}, \textbf{\textit{BDSD}}.

\subsubsection{Performances and Analyses}

The results of comparisons are summarized in Table \ref{tab:result3}. Some examples are also displayed in Figure~\ref{fig:comparison} to demonstrate the advantages of our proposed methods. As shown, our methods consistently outperform the compared methods across all metrics, demonstrating superior overall segmentation accuracy, better boundary alignment and precision, and a higher degree of shape preservation. 

\begin{table}[htb]
    \centering
  \caption{Comparison Experiments with the Most Related Methods}  \label{tab:result3}
  \begin{tabular}[c]{c|ccc|ccc}
  \hline
    \hline
        \multirow{2}{*}{\textbf{Method}}  & \multicolumn{3}{c|}{\textbf{BUSI}} & \multicolumn{3}{c}{\textbf{Kvasir}}\\ 
       & \textbf{\textit{Dice}}{$\uparrow$} & \textbf{\textit{BD}}{$\downarrow$} & \textbf{\textit{BDSD}}{$\downarrow$} & \textbf{\textit{Dice}}{$\uparrow$} & \textbf{\textit{BD}}{$\downarrow$} & \textbf{\textit{BDSD}}{$\downarrow$}  \\
       \hline
       UNET\citep{ronneberger2015u} & 70.38 & 18.12 & 16.28 & 66.19 & 20.44 & 16.72 \\
$+\loss_{perim}$\citep{jurdi2021surprisingly} & 73.52 & 15.23 & 9.789 & 70.52 & 16.74 & 15.48 \\
$+\loss_{AB}$\citep{wang2022active} & 74.17 & 15.19 & 10.64 & 69.21 & 15.57 & 14.74 \\
       $+\loss_{S}$ & \textbf{75.76} & \textbf{12.58} & \textbf{9.740} & \textbf{74.54} & \textbf{13.73} & \textbf{12.35} \\
       \hline
       \hline
       Dcan\citep{chen2016dcan}& 72.41 & 17.65 & 16.19 & 70.72 & 18.79 & 17.04 \\
       Dmtn\citep{tan2018deep} & 74.02 & 15.77 & 13.59 & 71.80 & 17.39 & 15.89 \\
       Psi-net\citep{murugesan2019psi} & 73.16 & 17.75 & 13.41 & 71.88 & 15.98 & 13.14 \\
       ConvMCD\citep{murugesan2019conv} & 74.92 & 13.98 & 9.839 & 74.21 & 15.27 & 14.76 \\
       \textbf{CFSSnet} & \textbf{76.29} & \textbf{11.08} & \textbf{8.510} & \textbf{79.62} & \textbf{12.38} & \textbf{12.25} \\
       \hline
       \hline
    \end{tabular}
\end{table}

\subsection{Summarized Analysis }

An important observation drawn from experiments is that the proposed shape loss $\mathcal{L}_{S}$ and corresponding variational iterations (Algorithm~\ref{algo1}) have the main impact and primarily drives improvements in boundary‐distance metrics (\textit{\textbf{BD}}, \textit{\textbf{BDSD}}) rather than overlap‐based \textit{\textbf{Dice}}. Such improvements are consistent across datasets, suggesting that the proposed shape constraint consistently enforces tighter adherence to the true contour across different segmentation tasks and imaging modalities. The reason for this discrepancy is that our shape constraint is derived from definition of global shape similarity and explicitly targets boundary alignment, so it directly penalizes deviations along boundary. By contrast, \textit{\textbf{Dice}} depends on the overall region‐overlap, which is indirectly improved by including the proposed shape constraint.  

\subsection{Ablation Study}

\subsubsection{Comparison for Shape Loss with Established Loss} 
As demonstrated in \citep{isensee2021nnu}, the most common loss function for medical segmentation is the sum of cross-entropy and Dice loss (DiceCE). Although in previous sections we have conducted extensive experiments on adding proposed shape loss onto cross-entropy loss and Dice loss alone, we are still interested in comparison of shape loss with this established loss function. We use U-net as baseline, and compare several different combination of loss functions on the BUSI and Kvasir datasets in Table~\ref{tab:abcedice}. One can observe dice loss in general yields better \textit{\textbf{Dice}} score, but by adding shape loss, shape metrics can still have improvement.

\begin{table}[htb]
    \centering
  \caption{Comparison on different combinations of shape loss and established loss}  \label{tab:abcedice}
  \begin{tabular}[c]{c|ccc|ccc}
  \hline
    \hline
        \multirow{2}{*}{\textbf{Loss}}  & \multicolumn{3}{c|}{\textbf{BUSI}} & \multicolumn{3}{c}{\textbf{Kvasir}}\\ 
       & \textbf{\textit{Dice}}$\uparrow$ & \textbf{\textit{BD}}$\downarrow$ & \textbf{\textit{BDSD}}$\downarrow$ & \textbf{\textit{Dice}}$\uparrow$ & \textbf{\textit{BD}}$\downarrow$ & \textbf{\textit{BDSD}}$\downarrow$  \\
       \hline
       $\loss_{DiceCE}$ & 75.18 & 13.47 & 10.74 & 76.46 & 13.06 & 11.89 \\
       $\loss_{CE}+\loss_{S}$ & 75.76 & \underline{12.58} & 9.740 & 74.54 & 13.73 & 12.35 \\
$\loss_{Dice}+\loss_{S}$ & 76.18 & 13.48 & 10.84 & \underline{76.83} & 12.86 & 11.67 \\
$\loss_{DiceCE} +\loss_{S}$ & \underline{76.69} & 13.04 & \underline{9.486} & 76.35 & \underline{12.26} & \underline{11.20} \\
       \hline
       \hline
    \end{tabular}
\end{table}

\subsubsection{ Contour Flow Branch and Shape Constraint Module in CFSSnet}
In this section, an ablation study is conducted to understand the individual contribution of modules in CFSSnet, i.e. the contour flow branch and shape constraint module. By doing so, we can get a better understanding of where the inherent shape-preserving ability of CFSSnet comes from.  To proceed, we modify $T$, the number of shape constraints modules from no such modules at all, and increase $10$ modules each time until $T=40$ to check the best choice of module number to implement. For a contribution of the contour flow branch, we do comparable experiments to see how metrics vary with and without the branch. If there is no such branch, $\F = [\cos (\textrm{conv}(U)), \sin (\textrm{conv}(U))]$, i.e. directly from backbone logits output. These different settings are trained and evaluated on the WBC2 dataset to see whether the inherent shape-preserving ability of CFSSnet is retained or not.

\textit{Why contour flow prediction branch is necessary?} As shown in Table~\ref{tab:resultablation}, the inclusion of the contour flow branch leads to a great improvement in the overall performance of CFSSnet. Without this branch, the network lacks the representational capacity to accurately predict the contour flow $\F$. Notably, even in the absence of shape constraint modules, the network benefits from learning the ground truth flow, which enhances shape similarity.

\textit{What is the optimum number of shape constraint modules?} Our ablation study suggests that using $T=20$ shape constraint modules offers the best overall performance, which is the setting for this paper's implementation. With this configuration, the network strikes an ideal balance between \textbf{\textit{Dice}}, shape properties (\textbf{\textit{BD}}, \textbf{\textit{BDSD}}), and resistance to noise. The reasoning is that our shape constraint modules are essentially an unrolled iterative process for solving the variational segmentation model, which requires a certain number of iterations to reach an optimal solution. When $T<20$, the iterations may be insufficient. However, when $T>20$, further iterations do not provide any additional benefit. Moreover, adding unnecessary modules can overcomplicate the gradient graph during backpropagation, making it harder for network convergence. We recommend using $T$ in the range of [15, 25].

\begin{table}[htb]
    \centering
  \caption{Ablation study: individual influence of Contour Flow(CF) Prediction branch and Shape Constraint (SC) Modules}
  \begin{tabular}[c]{cc|ccc}
  \hline
    \hline
       \multirow{2}{*}{\textbf{CF Prediction Branch}} & \multirow{2}{*}{\textbf{SC Modules}} & \multicolumn{3}{c}{\textbf{WBC2}}\\ 
       & & \textbf{\textit{Dice}}{$\uparrow$} & \textbf{\textit{BD}}{$\downarrow$} & \textbf{\textit{BDSD}}{$\downarrow$} \\
    \hline
    \hline
      \xmark & \multirow{2}{*}{\xmark} & 94.75 & 4.579 & 3.834 \\
      \cmark &  & \underline{95.92} & \underline{3.250} & \underline{2.917} \\
     \hline
      \multirow{5}{*}{\xmark} & $T=10$ & 93.57 & 4.112 & 2.958 \\
      & $T=20$ & 94.38 & 4.422 & 3.679 \\
      & $T=30$ & 94.08 & 4.461 & 3.878 \\
      & $T=40$ & 91.54 & 6.341 & 5.921 \\
     \hline
      \multirow{5}{*}{\cmark} & $T=10$ & \textbf{96.15} & 3.528 & 3.610 \\
       & $T=20$ & 95.68 & \textbf{3.118} & \textbf{2.839}\\
      & $T=30$ & 95.56 & 3.569 & 3.607\\
      & $T=40$ & 94.77 & 4.241 & 3.652\\
      \hline
      \hline
  \end{tabular}
  \label{tab:resultablation}
\end{table}

\subsubsection{Impact of Contour Flow Estimation Errors}

\begin{figure}
\centering
\includegraphics[width=0.7\linewidth]{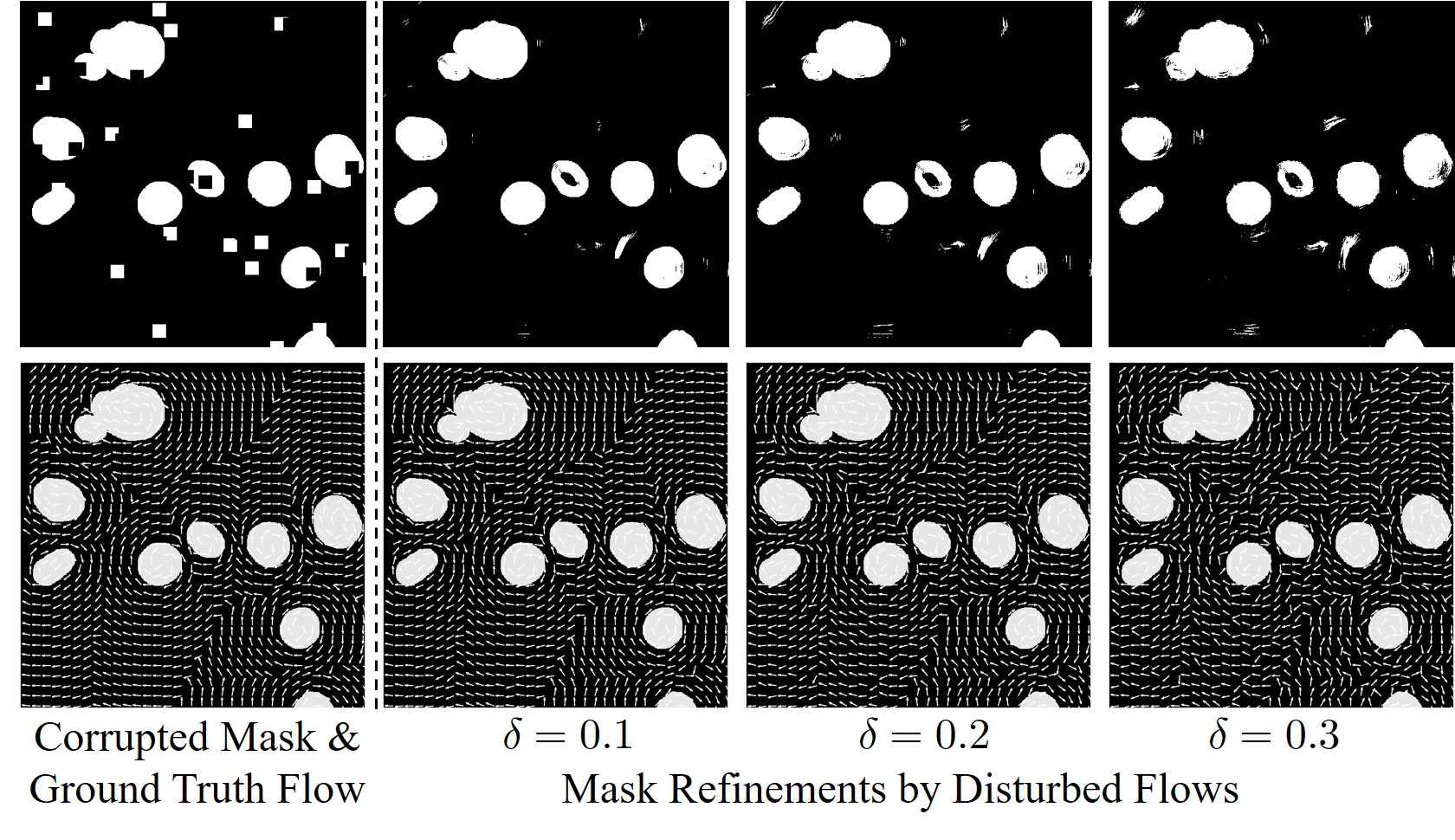}%
\caption{Mask Refinements by Algorithm~\ref{algo1} with disturbed flow.
}
\label{fig:toy2}
\end{figure}

In practice, the learned contour flow \(\flow\) may differ from the ground truth due to estimation errors. To study how such errors affect iterative refinement in Algorithm~1, we first present a controlled toy experiment. Let $\hat{\F} \;=\; \flow \;+\; e, 
  \quad e \sim \mathcal{N}(0,\;\delta^2 I)$, where \(\delta\) controls the noise level. We apply Algorithm~\ref{algo1} on a patch corrupted toy image $I_2 = I + p$. (the same “damage” procedure used in Section~\ref{sec:CFSS}) using \(\hat{\F}\) instead of the exact \(\flow\). Specifically, we test \(\delta \in \{0.1,\,0.2,\,0.3\}\). The resulting segmentation masks are shown in Figure~\ref{fig:toy2}. Even when \(\delta=0.3\), a fairly large disturbed flow, Algorithm~1 still manages to produces a better refinement. This demonstrates that Algorithm~\ref{algo1} is robust to moderate levels of error in \(\flow\).

Next, we evaluate the magnitude of flow estimation errors on the Kvasir datasets on both training and inference phases and compare two configurations: (1) our standard CF branch, equal depth to the segmentation branch, and (2) an enlarged CF branch with double depth. For each test image, we compute the predicted flow \(\F\) and compare it to the ground truth flow \(\flow\) using three metrics: Average Cosine Similarity (ACS), End‐Point Error (EPE),  and Average Divergence Error (ADE). Table~\ref{tab:abcf} summarizes these metrics on both phases for the two CF‐branch configurations. We observe that while some error in learning \(\flow\) is unavoidable, current CF‐branch depth has moderate estimation errors in \(\flow\). Using deeper CF branch has marginal improvement, but will add cost of computations.

\begin{table}[htb]
    \centering
  \caption{Contour Flow Learning Errors on Kvasir Dataset}  \label{tab:abcf} 
  \begin{tabular}[c]{c|ccc|ccc}
  \hline
    \hline
        \multirow{2}{*}{\textbf{Depth}}  & \multicolumn{3}{c|}{\textbf{Training}} & \multicolumn{3}{c}{\textbf{Inference}}\\ 
       & \textbf{\textit{ACS}}$\uparrow$ & \textbf{\textit{EPE}}$\downarrow$ & \textbf{\textit{ADE}}$\downarrow$ & \textbf{\textit{ACS}}$\uparrow$ & \textbf{\textit{EPE}}$\downarrow$ & \textbf{\textit{ADE}}$\downarrow$  \\
       \hline
       Equal & \underline{0.9701} & \underline{0.1525} & 0.0391 & 0.8975 & 0.2885 & 0.0436 \\
Double & 0.9621 & 0.1801 & \underline{0.0384} & \underline{0.9050} & \underline{0.2831} & \underline{0.0421} \\
       \hline
       \hline
    \end{tabular}
\end{table}

\section{Conclusion and Discussion}

In this paper, we explore a global shape similarity between the segmentation function and the ground truth, defined by the two sharing similar contours. We mathematically derive an equivalent contour flow constraint condition to achieve this similarity. More importantly, we successfully implement the constraint into deep neural networks by proposing the shape loss term and the unrolled CFSSnet. Experiments showcase the adaptability of the shape loss to enhance the performance of arbitrary learning-based segmentation networks, and the inherent ability of the CFSSnet to preserve the global similarity.

Despite the demonstrated strengths of our shape loss and CFSSnet framework, several limitations remain. First, incorporating the shape loss introduces additional computational overhead during both training and inference, which may hinder scalability on very large volumes or in resource-constrained settings. Second, the method could be sensitive to class imbalance: rare or small structures may not receive sufficient emphasis in the shape prior, leading to suboptimal segmentation for imbalanced classes. Third, while our approach effectively enforces global shape similarity, it does not theoretically guarantee the preservation of complex topological properties such as connectivity or genus. In future work, we plan to investigate more efficient optimization strategies to reduce computation time, develop adaptive weighting schemes to mitigate class imbalance, and integrate explicit topological constraints or priors to ensure connectivity and other critical topological invariants.

Although we have only discussed the preservation of binary classification object shapes, the same idea can be applied to multi-classification. Multi-class segmentation tasks can always be decomposed
into some binary sub-problems regarding each class to implement our method. One
can also extend the proposed contour flow with a softmax manner to be class-wise constrained.

\section*{Supplementary Material}

\subsection*{Proof of Corollary 1}
\begin{proof}
    $\partial\Omega_ = \{ i\in\Omega| \segu(i) = \gamma\in\R\}$, and $\conu^{\gamma,m}$ is the $m$-th $\gamma$-contour branch of $\segu$. Since $\segu$ has contour similarity with $\gt$, by definition, $\exists \con^{\alpha,n}$ such that $\conu^{\gamma,m} = \con^{\alpha,n}$. By definition of $\gt$'s contours, $\forall i,j\in \conu^{\gamma,m}, \sdf(i) = \sdf(j) = \alpha$. Therefore, $D(i,\partial\gtsp) = |\sdf(i)| = |\alpha| = |\sdf(j)| = D(j,\partial\gtsp).$
\end{proof}

\subsection*{Proof of Lemma 1}
\label{append1}
For $f\in C^1(\Omega)$, consider $\nabla f(i)\in\R^2$ and $\alpha$-contour $\conf^{\alpha}=\{ j\in\Omega | f(j)=\alpha \}$, where $\alpha = f(i)$. If $\conf^{\alpha}$ is complex-connected curve with several connected components, we can simply focus on the one branch where $i$ is located. Therefore, we assume $\conf^{\alpha}$ is a $C^1$ Jordan curve, which can be represented by $f(j)\equiv\alpha$.  Introduce an arc-length parameter $t\in\R$ to represent  it as a parametric curve $\varphi(t) = j\in \Omega$. Then, by derivative of $t$ at both sides of $f(j)\equiv\alpha$:
\begin{equation}
    \frac{d}{dt}f(j) = \frac{d}{dt}f(\varphi(t)) = \nabla f(\varphi(t)) \cdot \varphi^{\prime} (t)\equiv 0 = \frac{d}{dt} \alpha,
\end{equation}
where $\nabla f(\varphi(t)) =\nabla f(j)$ is the gradient vector of $f$ at $j$, and $\varphi^{\prime} (t)$ is the tangent vector of contour branch $\conf^{\alpha}$ at $j$. This equation indicates the two are orthogonal with each other at any locations on the $\alpha$-contour, i.e. $\nabla f(i) \perp \conf^{\alpha}$.

\subsection*{Proof of Theorem 1}
\label{append2}

Recall contour similarity $\conu\sim\con$ means $\forall \conu^{\beta, m}$, i.e. arbitrary contour branch of $\segu$, there exists $\con^{\alpha, n}$, i.e. contour branch of $\gt$, such that $\conu^{\beta, m} = \con^{\alpha, n}$. Since $\conu^{\beta, m}, \con^{\alpha, n}$ are $C^1$ Jordan curve, we can always introduce an arc-length parameter $t\in\R$ to represent a contour branch by parametric curve $\varphi(t)=i, \mathbbm{R}\mapsto\con^{\alpha,n}$, where the arc-length parameter $t$ increases along the tangent direction $\flow(i)=\varphi^{\prime}(t)$. 

($\Longrightarrow$:) If $\conu\sim\con$, then $\forall i\in\Omega$, denote $\beta = \segu(i)$, $i\in \conu^{\beta, m}$ (some branch of $\conu^{\beta}$). By definition of contour similarity, $\exists \con^{\alpha, n}$, such that $\conu^{\beta, m} = \con^{\alpha, n}$. Therefore, $\segu(i)\equiv \beta, i\in \con^{\alpha, n}$. Now, use the parametric form to represent $\con^{\alpha,n}$ by $\varphi(t)=i\in \con^{\alpha,n}$. By derivative of $t$ at both sides of $\segu(i)\equiv \beta$:
\begin{equation}
    \frac{d}{dt}\segu(i) = \frac{d}{dt}\segu(\varphi(t)) = \nabla \segu(\varphi(t)) \cdot \varphi^{\prime} (t) = 0 = \frac{d}{dt} \beta,
\end{equation}
where $\nabla \segu(\varphi(t)) = \nabla \segu(i), \varphi^{\prime}(t) = \flow(i)$. To conclude, $\forall i\in\Omega, \nabla \segu(i)\cdot \flow(i)=0$.

($\Longleftarrow$:) We show it by contradiction. If $\exists \beta^* \in \mathbbm{R}, m^*\le M, \conu^{\beta^*, m^*}$ such that $\forall \con^{\alpha,n}, \conu^{\beta^*, m^*}\neq \con^{\alpha,n}$, then we know $\conu^{\beta^*, m^*}$ is not possible a contour of $\gt$. Parameterize $\conu^{\beta^*, m^*}$ by $\psi(t) = i\in\conu^{\beta^*, m^*}$. By definition of contour, $\sdf(i) = \sdf(\psi(t)) \coloneqq H(t) \neq {\rm const}, i\in \conu^{\beta^*, m^*}$. Therefore, $\exists t^*$, such that $H^{\prime}(t^*) \neq 0$, otherwise $H$ is constant for $\forall t$. 

We denote $i^* = \psi(t^*)\in\conu^{\beta^*, m^*}$. Then, $H^{\prime}(t^*) = \nabla \sdf(i^*) \cdot \psi^{\prime}(t^*)\neq 0$. Here, $\psi^{\prime}(t^*)$ is the tangent vector of $\conu^{\beta^*, m^*}$ at $i^*$. According to Lemma 1, $\psi^{\prime}(t^*)$ is orthogonal with $\nabla\segu(i^*)$. Moreover, $\nabla\sdf(i^*)$ is also orthogonal with $\flow(i^*)$ by Definition 4. Thus, by rotating vectors $\nabla\sdf(i^*), \psi^{\prime}(t^*)$ with $\pi/2$, we get $\nabla \segu(i^*)\cdot \flow(i^*)\neq0$, which is contradicted with $\forall i\in\Omega, \nabla \segu(i)\cdot \flow(i)=0$.

\subsection*{Proof of Theorem 2}
\label{append3}
Since
\begin{equation*}
    \begin{array}{rl}
  \displaystyle\biggl.\int_{\Omega}  \segu~{\rm div}(q\flow)  di&=-\displaystyle\biggl.\int_{\Omega}   q(\nabla\segu\cdot \flow)  di+\displaystyle\biggl.\int_{\partial\Omega}  \segu q \flow\cdot\vec{\boldsymbol{n}} dS\\
  &=-\displaystyle\biggl.\int_{\Omega}   q(\nabla\segu\cdot \flow)  di,
  \end{array}
\end{equation*}
then $\displaystyle\biggl.\max_{q\in C^1(\Omega)} \left\{\int_{\Omega}  \segu~{\rm div}(q\flow)  di\right\}=0$ if $\segu\in\mathbb{C}$. Else, taking $q=k(\nabla\segu\cdot \flow)\in C^1(\Omega)$ and let $k$ go to $+\infty$, then one can obtain  $\displaystyle\biggl.\max_{q\in C^1(\Omega)} \left\{\int_{\Omega}  \segu~{\rm div}(q\flow)  di\right\}=+\infty.$ which completes the proof.

\end{document}